\documentclass[ba,preprint]{imsart}% use this for supplement article
\RequirePackage{amsthm,amsmath,amsfonts,amssymb}
\RequirePackage[numbers]{natbib}
\RequirePackage{bm}
\RequirePackage{float}
\RequirePackage{algorithm}
\RequirePackage{algorithmic}
\RequirePackage[colorlinks,citecolor=blue,urlcolor=blue,backref=page]{hyperref}
\RequirePackage{graphicx}
\usepackage{adjustbox}
\usepackage{subcaption}  % replacement for subfigure
\usepackage{booktabs}
\usepackage{multirow}
\usepackage[autostyle, english = american]{csquotes}
\MakeOuterQuote{"}
\graphicspath{{Figures/}}
\pubyear{2026}
\arxiv{2026.00000}
\volume{TBA}
\issue{TBA}
\firstpage{1}
\lastpage{1}

\startlocaldefs
\theoremstyle{plain}

\newtheorem{theorem}{Theorem}[section]

\newtheorem{proposition}[theorem]{Proposition}
\theoremstyle{definition}

\theoremstyle{remark}

\endlocaldefs

\begin{document}
	
	\begin{frontmatter}
		\title{A Bayesian Boolean Matrix Factorization with Application to Copy Number Analysis in Cancer}
		%\title{A sample article title with some additional note\thanksref{t1}}
		%\runtitle{A sample running head title}
		%\thankstext{T1}{A sample additional note to the title.}
		
		\begin{aug}
			%%%%%%%%%%%%%%%%%%%%%%%%%%%%%%%%%%%%%%%%%%%%%%%
			%% Authors                                  %%
			%%%%%%%%%%%%%%%%%%%%%%%%%%%%%%%%%%%%%%%%%%%%%%%
			\author[A,B]{\fnms{Adolphus }~\snm{Wagala}}
			\author[A,B]{\fnms{Mehmet}~\snm{Samur}}
			\and
			\author[A,B]{\fnms{Giovanni}~\snm{Parmigiani}}
			
			%%%%%%%%%%%%%%%%%%%%%%%%%%%%%%%%%%%%%%%%%%%%%%%
			%% Addresses                               %%
			%%%%%%%%%%%%%%%%%%%%%%%%%%%%%%%%%%%%%%%%%%%%%%%
			\address[A]{Department of Data Science, Dana-Farber Cancer Institute}
			
			\address[B]{Department of Biostatistics, Harvard T.H. Chan School of Public Health}
			
			\runauthor{Wagala, Samur and Parmigiani}
		\end{aug}
		
		\begin{abstract}
        Binary data factorization arises in many fields. Standard factorization methods based on real-valued arithmetic often fail to respect the discrete structure or yield interpretable decompositions. Boolean Matrix Factorization (\texttt{BooMF}) addresses this by decomposing a binary matrix into two lower-rank matrices using logical \texttt{AND} and \texttt{OR} instead of real-valued operations, approximately representing the data as a Boolean disjunction of interpretable binary patterns. In cancer genomics, unlike rotational or additive decompositions, \texttt{BooMF} can reveal coordinated changes in multiple features that may drive tumor evolution. Most existing \texttt{BooMF} methods are heuristic and greedy, making them sensitive to initialization, prone to local optima, and lacking principled model selection or uncertainty quantification. We address these issues with a Bayesian Boolean Matrix Factorization (\texttt{BBMF}) model defined by a generative process and sparsity-inducing priors. The model enforces Boolean constraints, yields interpretable latent factors with coherent uncertainty quantification, and is fully conjugate, so all full conditionals have closed-form expressions, enabling Gibbs sampling without numerical approximations. Cancer evolution can involve widespread, near-simultaneous chromosome-number changes (e.g., whole-genome duplication followed by brief genomic instability and strong selection), which Boolean factorizations capture more accurately and interpretably than additive models. In simulation experiments, we compare \texttt{BBMF} with the widely used \texttt{Asso} and \texttt{GreConD+} boolean factorization algorithms. In all the simulation experiments,  \texttt{BBMF} more reliably recovers the true latent factors—seen in strongly diagonally dominant similarity matrices and factor patterns closely matching ground truth, and is competitive with or superior to alternatives in reconstruction accuracy and standard classification metrics. We applied \texttt{BBMF} to copy-number alteration data at the level of chromosomal arms in multiple myeloma, where binary entries indicate the presence or absence of arm-level amplifications. The model identifies a small number of interpretable bicliques linking subsets of patients to recurrently co-altered chromosomal arms, yielding a compact, biologically meaningful summary of tumor heterogeneity and demonstrating its utility for discovering discrete latent structure in complex binary data.
		\end{abstract}
		
		
		
	\end{frontmatter}
	\newpage
	\section{Introduction}
	\label{sec:Introduction}

    \subsection{Context}
Binary data arise naturally across a wide range of domains. A fundamental challenge in analyzing such data is identifying compact and interpretable structures that faithfully capture underlying patterns.

Boolean Matrix Factorization \texttt{(BooMF)} addresses this challenge by approximating an observed binary matrix $\mathbf{X} \in \{0,1\}^{K \times G}$ as the Boolean product of two lower-dimensional binary matrices $\mathbf{W} \in \{0,1\}^{K \times R}$ and $\mathbf{H} \in \{0,1\}^{R \times G}$, where $R \ll \min(K,G)$ denotes the rank of factorization. The Boolean product is not over a field, as in more standard approaches, but over the Boolean semiring $(\{0,1\}, \lor, \land)$; therefore, these factorizations have some unique properties, as a result of which they can produce, in suitable settings, smaller reconstruction errors than factorizations of the same rank conducted under standard arithmetic \cite{miettinen2020recent}. The primary objective of \texttt{BooMF} is to discover latent variables, sometimes known as factors, that offer a more interpretable perspective on the input data set \cite{trnecka2022Boolean}. Several algorithms for \texttt{BooMF} have been presented in the literature. Most contemporary \texttt{BooMF} methods minimize the so-called reconstruction error, i.e., the number of input entries incorrectly predicted by the reconstruction \cite{trnecka2022Boolean}. A classical \texttt{BooMF} methodology is the \texttt{Asso} factorization algorithm based on the Discrete Basis Problem (DBP) proposed by \citet{Asso_4479462}. Another important \texttt{BooMF} approach is the approximate factorization problem (AFP) implemented in algorithms such as \texttt{GreConD} and \texttt{GreConD+} by \citet{Belohlavek2010jcss}. \citet{Belohlavek2010jcss} prove an optimality theorem that shows that the decompositions with the least number $R$ of factors are those in which the factors are formal concepts in the sense of formal concept analysis~\cite{ganter1999formal}. From the theorem, the set of formal concepts associated with the input matrix provides a non-redundant space of factors for optimal decomposition. At each step, \texttt{GreConD} greedily selects the formal concept that covers the largest number of as-yet-uncovered 1-entries of the input matrix, and terminates when all 1-entries have been covered or, if a user-specified cap $R$ on the number of factors is provided, after $R$ concepts have been added.
	
\texttt{BooMF} has also been implemented using probabilistic models. \citet{Rukat2017pmlr} proposed a probabilistic generative framework for Boolean Matrix Factorization, termed \texttt{OrMachine}, and developed a Metropolized Gibbs sampling scheme for posterior inference. \texttt{OrMachine} is a fully Boolean generative model: the factor matrices and their product remain binary throughout inference, and a single global dispersion parameter $\lambda \in \mathbb{R}_+$ controls the per-bit noise level, with the probability that each observation agrees with the deterministic Boolean prediction given by the logistic sigmoid $\sigma(\lambda)$. Because the method provides full posterior inference, posterior means can be thresholded to control false-positive rates, and the inferred patterns are more interpretable than those of point-estimate baselines. Related probabilistic treatments of BooMF include the multi-assignment clustering model of \citet{streich2009multi}, which uses a probabilistic generative process with a global noise source and infers point estimates by deterministic annealing; the non-parametric Bayesian model of \citet{wood2006non}, which uses an Indian Buffet Process prior to infer an unbounded number of latent binary causes via Gibbs sampling; and the probabilistic graphical model of \citet{ravanbakhsh2016boolean}, which derives a message-passing algorithm for MAP estimation and also supports noisy matrix completion.

It is useful to contrast \texttt{(BooMF)} with Binary Matrix Factorization (BMF), a class of low-rank matrix factorization methods for finding structure in binary data using standard algebraic inner products. Several BMF approaches have been proposed. Some retain a real-valued low-rank model and map its entries to $[0,1]$ via a logistic link, interpreting the linear predictor as the logit of a Bernoulli mean; examples include logistic PCA and binary PCA. In such cases, the factors and weights (the scores and loadings of the PCA-style decomposition) are not themselves constrained to be binary, while the logistic link constrains the predicted Bernoulli means to lie in $[0,1]$ \cite{farias2023generalized,lumbreras2020bayesian}. Other approaches relax the binary constraint to obtain a continuous optimization problem --- typically minimizing a reconstruction error such as the mean squared error (MSE) --- and recover binary factor matrices through penalty terms in the objective or by thresholding the continuous solution. \citet{zhang2010binary} introduced one such method, extending non-negative matrix factorization (NMF) to binary outputs and applying it to gene-expression biclustering. They also argue that discretizing continuous inputs prior to factorization need not lose information and can in fact improve robustness to noise. \citet{schachtner2009binary} similarly applied binary NMF---using alternating least squares with non-negativity constraints---to identify systematic failure patterns in semiconductor wafer test data. Across these BMF variants, the factorization operates over the reals, yielding continuous entries that approximate $0$ and $1$ rather than exact binary $(0/1)$ values. This is often undesirable in genuinely Boolean applications, where one typically expects interpretable on/off decisions or set memberships. Consequently, the presence of non-integer values can make the factors harder to interpret and less aligned with the underlying discrete structure.

    \subsection{Chromosomal Alterations in Multiple Myeloma}
	
    Multiple Myeloma (MM) is a malignancy of plasma cells---terminally differentiated B cells of post-germinal-center origin---usually driven by primary and secondary chromosomal events~\cite{clarke2024chromosomal, mikulasova2022chromosomal}. DNA alterations in MM include  chromosomal translocations, copy number abnormalities (CNAs), and point mutations. CNAs play an important role in MM pathogenesis: some influence prognosis and therapeutic decisions. Frequent CNA events include gains and losses of complete chromosomes or chromosome arms, such as deletions of 1p and gain of the long arm of chromosome 1 (1q)~\cite{cardona2021genetic,lannes2023multiple}. Two large-scale genomic events that drive MM evolution are particularly important. Whole-genome duplication~\cite{Bielski2018} doubles chromosome content, promoting further short-term chromosomal instability; chromothripsis~\cite{CortesCiriano2020} causes massive chromosome shattering and reassembly in a single crisis, generating complex rearrangements, oncogene amplification, and tumor suppressor loss in one step. These and other "catastrophic" events can define clinically relevant subgroups in MM~\cite{Maura:2019vt,Maura2020}.

    In MM, chromosomal alterations are frequently encoded as binary variables denoting whether specific events are present or absent in each patient. \texttt{BooMF} is well suited to such data: it identifies discrete latent patterns, including subgroups of patients defined by recurrent constellations of chromosomal events and alteration signatures that may correspond to important steps in tumor evolution. Despite this theoretical fit, \texttt{BooMF} remains underexplored for chromosomal alteration data. Beyond factorization-based methods, graph- and network-search approaches have also been used to discover functional modules from cancer mutation data; for instance, \citet{miller2011discovering} build a gene interaction network and identify significant modules with recurrent or mutually exclusive alterations using an algorithmic significance test.

    Our motivating observation is that these chromosomal alterations occur in discrete, coordinated patterns---groups of arms recurrently gained or lost together---a structure that Boolean logic captures directly and that continuous, additive decompositions fundamentally cannot. Although developed in the context of MM, the proposed model applies broadly to any binary matrix whose latent structure is well captured by a Boolean disjunction of binary factor patterns.
    
	\subsection{Motivating Example}
\begin{figure}[htbp]
		\centering
		\includegraphics[width=\textwidth]{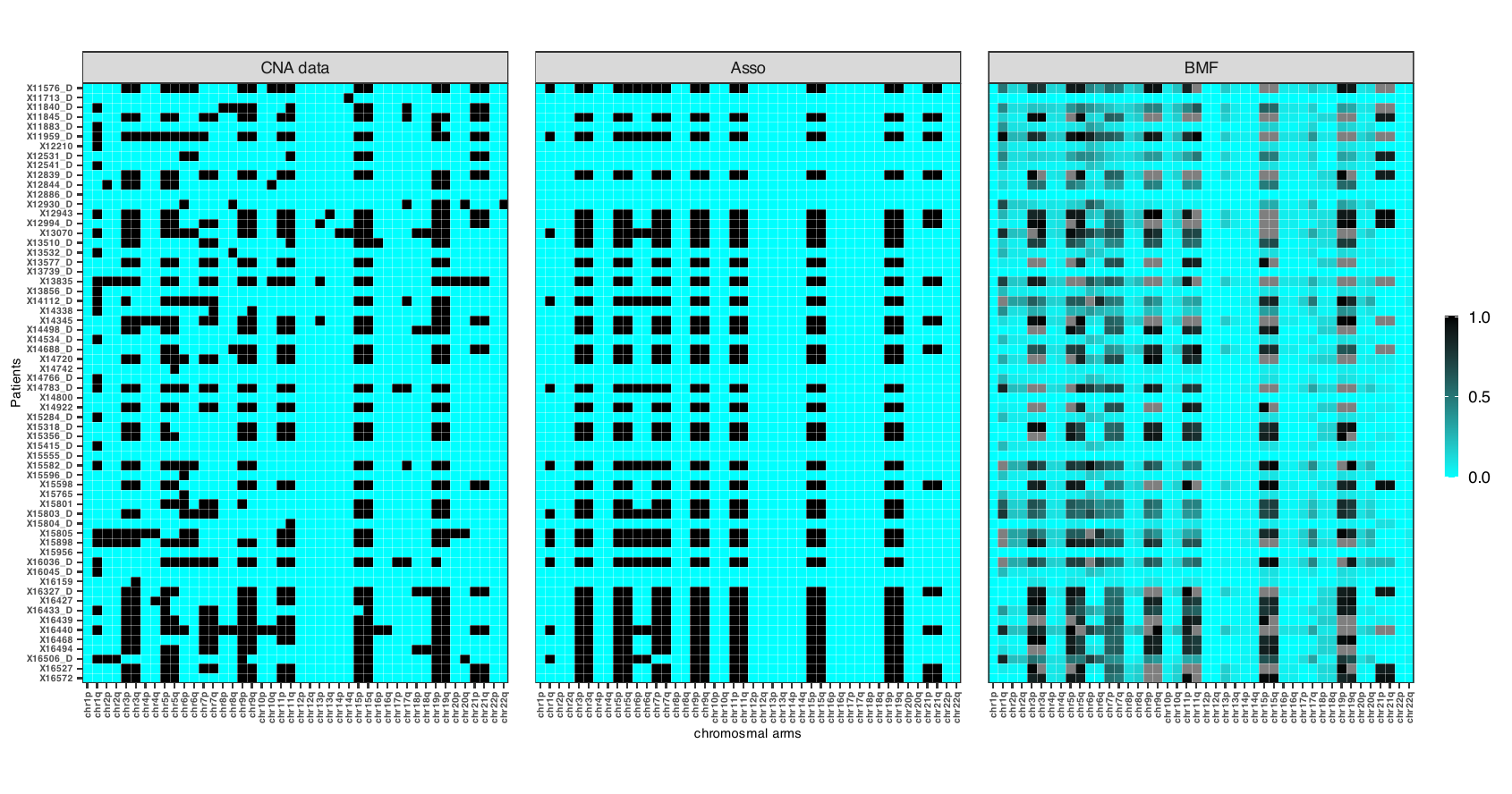}
		\caption{Original data (left) and reconstructed data sets from Asso (center) and BMF (right). 
        The rows represent individual patients and columns represent chromosomal arms. Each cell is encoded as 1 (black) for amplification and 0 (cyan) for non-amplification.}
		\label{fig:motivation_figures}
	\end{figure}

	To motivate the development of an alternative factorization algorithm, we first apply one method from each established family---\texttt{Asso} (a Boolean matrix factorization) and \texttt{BMF} (a real-valued factorization with binary constraints)---to the CNA data for multiple myeloma, derive the corresponding factor matrices, and then reconstruct approximate representations of the original matrix. The data matrix has $G = 44$ columns indexing chromosomal arms (the p and q arms of chromosomes 1--22) and $K$ rows indexing patients. In normal cells, each chromosomal arm is present in two copies, while in MM cells there can be more copies (an amplification) or fewer (a deletion). Let $\mathbf{X}\in \{0,1\}^{K\times G}$ be the input matrix, with $X_{kg}=1$ if arm $g$ is amplified in patient $k$ and $X_{kg}=0$ otherwise (normal copy number or deletion). Figure~\ref{fig:motivation_figures} shows the CNA data alongside the reconstructions produced by \texttt{Asso} and \texttt{BMF}. 
 
    The CNA matrix in Figure~\ref{fig:motivation_figures} exhibits a highly structured organization driven by two principal features: co-occurring alterations across chromosomal arms and patient subgroups characterized by shared CNA signatures. Co-occurring alterations manifest as block-like or parallel vertical patterns, indicative of recurrent joint amplifications rather than independent events. Such co-occurrence patterns are well documented in MM: for example \citet{sawyer2019acquired} describe ``jumping'' 1q translocations accompanied by concurrent deletions in receptor chromosomes; \citet{hanamura2021gain} report 1q21 amplification arising via tandem duplication and whole-arm translocations frequently accompanied by additional cytogenetic abnormalities; and \citet{Maura:2019vt} document recurrent co-occurrence of 1q gain and 13q loss at the population level. Within the matrix, horizontal bands denote loci recurrently amplified across the cohort, while rectangular blocks of rows with similar CNA profiles delineate patient subgroups with characteristic alteration signatures. These subgroups likely reflect underlying biological heterogeneity and distinct evolutionary trajectories, potentially corresponding to established cytogenetic categories, prognostic strata, or differential therapeutic responses; superimposed scattered single-patient gains represent additional private alterations. Together, this mosaic of co-amplified blocks, recurrent bands, and sporadic events suggests latent CNA factors---sets of chromosomal arms that repeatedly co-occur across patients---motivating the use of matrix factorization or related dimensionality-reduction methods to reveal shared structure, reduce data complexity, and identify recurrent CNA signatures with potential biological and clinical relevance.

    The choice of factorization method, however, critically determines whether the recovered structure faithfully respects the binary nature of the data. As Figure~\ref{fig:motivation_figures} illustrates, \texttt{Asso} produces strictly binary, blocky reconstructions consistent with the Boolean nature of the data, but is heuristic, deterministic, and fails to account for noise, uncertainty, or ambiguous patterns. In contrast, the \texttt{BMF} variant applied here produces continuous-valued reconstructions, as seen in the graded color scales. These considerations motivate a Bayesian Boolean Matrix Factorization (\texttt{BBMF}), which preserves binary structure while quantifying uncertainty and incorporating prior knowledge.
	
	%%%%%%%%%%%%%%%%%%%%%%%%%%%%%%%%%%%%%
	\subsection{Our Contributions}
	The proposed Bayesian Boolean Matrix Factorization (\texttt{BBMF}) model extends prior probabilistic \texttt{BooMF} in three concrete ways.

    First, the likelihood adopts an asymmetric two-parameter noise model, with separate sensitivity ($p_{11}$) and false-positive ($p_{10}$) rates. The closest probabilistic model, \texttt{OrMachine} \citep{Rukat2017pmlr}, parameterizes noise through a single parameter that constrains the two error types to be equal; \texttt{BBMF}'s noise model is a generalization, appropriate when false positives and false negatives occur at different rates --- as is often the case for binarized genomic measurements.

    Second, \texttt{BBMF} places hierarchical Beta priors on the per-sample and per-feature factor-activation probabilities, with a continuous spike-and-slab prior on the per-feature activation rates. The prior lets the data identify features that participate in essentially no factor; because such features can suppress entries across entire factor rows, the effective rank of the recovered factorization can be smaller than the user-supplied upper bound $R$. \texttt{OrMachine}'s fixed independent Bernoulli priors do not adapt in this way.

    Third, we introduce an aggregate factor-alignment similarity (AFAS) diagnostic for monitoring factor-recovery stability across MCMC iterations. AFAS complements standard reconstruction-error summaries by tracking how faithfully the latent structure --- and not just the reconstruction --- is recovered at each iteration.

    Because all full conditionals are closed-form Bernoulli or Beta updates, posterior inference proceeds by plain Gibbs sampling without numerical approximations. We illustrate \texttt{BBMF} on two simulation scenarios and benchmark it against the established \texttt{Asso} and \texttt{GreConD+} algorithms; we then apply it to arm-level copy-number alteration profiles in multiple myeloma, where the recovered bicliques align with the previously recognized hyperdiploid signature.

    The remainder of this paper is structured as follows. Section~\ref{sec:proposed_BBMF} introduces the proposed \texttt{BBMF} model. Section~\ref{sec:diagnostics} outlines the evaluation criteria and diagnostic procedures. Section~\ref{sec:sim_Expts} presents the simulation experiments. Section~\ref{sec:applic_real_data} demonstrates the application of the method to multiple-myeloma copy-number data. Section~\ref{sec:summary_conclusions} summarizes the main findings and offers concluding remarks.
	
	\section{A Bayesian Boolean Matrix Factorization (BBMF) Model}
	\label{sec:proposed_BBMF}
	\subsection{Notation and preliminaries}
    Throughout this paper, matrices are denoted by bold uppercase letters. Recalling the matrix $\mathbf{X} \in \{0,1\}^{K \times G}$ introduced in Section~\ref{sec:Introduction}, we seek a factorization of $\mathbf{X}$ as the Boolean product of two binary matrices $\mathbf{W} \in \{0,1\}^{K \times R}$ and $\mathbf{H} \in \{0,1\}^{R \times G}$, where $R \in \mathbb{Z}_+$ is the factorization rank with $R \ll \min(K,G)$. Row and column vectors extracted from these matrices are denoted by the same bold uppercase letter with a dot subscript indicating orientation: $\mathbf{W}_{k\cdot}$, $k=1,\ldots,K$, denotes the $k^{th}$ row of $\mathbf{W}$, and $\mathbf{H}_{\cdot g}$, $g=1,\ldots,G$, denotes the $g^{th}$ column of $\mathbf{H}$. The factorization takes the form
	\begin{align}
		\mathbf{Z}=\mathbf{W} \circ \mathbf{H}
		= \left[ \bigvee_{r=1}^R \left( W_{kr} \wedge H_{rg} \right) \right]_{k=1,\dots,K;\;g=1,\dots,G}
		\label{Eq:1}
	\end{align}
	where $\circ$ denotes the Boolean matrix product, which is analogous to standard matrix multiplication except that the multiplication is replaced by logical \texttt{AND} ($\land$) and addition by logical \texttt{OR} ($\lor$); for example, $1 \times 1 = 1 \land 1 = 1$ and $1 + 1 = 1 \lor 1 = 1$.
	
	\subsection{The Proposed Model}
    Let $X_{kg}$ denote the observed binary entry at row $k$ and column $g$ of $\mathbf{X}$, and let $Z_{kg}$ be the corresponding estimated entry of $\mathbf{Z}$. Assuming that the observation error rates do not vary with the sample $k$ or feature $g$, but differ depending on the type of error, then formally,
	\begin{align*}
		%\begin{split}
		P(X_{kg}=1 \mid Z_{kg}=1) = p_{11}, & \quad
		P(X_{kg}=0 \mid Z_{kg}=1) = 1-p_{11}, \\
		P(X_{kg}=1 \mid Z_{kg}=0) = p_{10}, &\quad
		P(X_{kg}=0 \mid Z_{kg}=0) = 1-p_{10}.
		%\end{split}
	\end{align*}
	Here, $p_{11}$ denotes the sensitivity (true positive rate), while $p_{10}$ denotes the false positive rate. Together, these parameters govern the misclassification process for all $(k,g)$. The generative process for each entry $X_{kg}$ is
	\begin{align*}
		P(X_{kg}\mid Z_{kg})
		&= \left[p_{11}^{X_{kg}} (1-p_{11})^{1-X_{kg}}\right]^{Z_{kg}}
		\times
		\left[p_{10}^{X_{kg}} (1-p_{10})^{1-X_{kg}}\right]^{1-Z_{kg}}.
	\end{align*}
	The following prior distributions are imposed on $W_{kr}$ and $H_{rg}$:
	\begin{align*}
		W_{kr}\mid \alpha_k &\stackrel{\text{i.i.d.}}{\sim} \operatorname{Bernoulli}(\alpha_k), \quad r=1,\dots,R,\ k=1,\dots,K,\\
		H_{rg}\mid \beta_g &\stackrel{\text{i.i.d.}}{\sim} \operatorname{Bernoulli}(\beta_g), \quad r=1,\dots,R,\ g=1,\dots,G,
	\end{align*}
	where the i.i.d.\ statement holds across $r$ for each fixed $k$ (respectively $g$).
    %--------------------------------------------------------------------------------------
    To induce sparsity in the factor matrix $\mathbf{H}$, we introduce a Bernoulli latent variable $\psi_g$ and specify a two-component (continuous spike-and-slab) prior on $\beta_g$:
    \begin{align*}
    \beta_g \mid \psi_g \sim
    \begin{cases}
    \text{Beta}(b_1, b_2), & \psi_g = 1, \\
    \text{Beta}(c_1, c_2), & \psi_g = 0,
    \end{cases}
    \qquad \text{with} \quad
    \psi_g \sim \text{Bernoulli}(\pi).
    \end{align*}
    %-------------------------------------------
Furthermore, we place the following priors on the remaining hyperparameters,
	\begin{align*}
	\alpha_k &\sim \text{Beta}(a_1,a_2),\\
	\pi &\sim \text{Beta}(d_1,d_2).
	\end{align*}
	These specifications imply the standard conditional independence structure of the model: the observed entries $X_{kg}$ are conditionally independent across $(k,g)$ given $(\mathbf{W}, \mathbf{H})$, and the remaining variables are a priori independent except where connected by the prior factorization above. The resulting posterior distribution is
	\begin{align}
        & P(\mathbf{W},\mathbf{H},\boldsymbol{\alpha},\boldsymbol{\beta},\boldsymbol{\psi},\pi \mid \mathbf{X})
        \propto P(\mathbf{X} \mid \mathbf{W},\mathbf{H})
        \times \prod_{k=1}^{K} P(\mathbf{W}_{k\cdot} \mid \alpha_k)
        \times \prod_{g=1}^{G} P(\mathbf{H}_{\cdot g} \mid \beta_g) \notag \\
        & \quad \times \prod_{k=1}^{K} P(\alpha_k)
        \times \prod_{g=1}^{G} P(\beta_g \mid \psi_g)
        \times \prod_{g=1}^{G} P(\psi_g \mid \pi)
        \times P(\pi) \notag \\
        & \propto \prod_{k=1}^{K} \prod_{g=1}^{G} \left[p_{11}^{X_{kg}} (1-p_{11})^{1-X_{kg}}\right]^{Z_{kg}} \left[p_{10}^{X_{kg}} (1-p_{10})^{1-X_{kg}}\right]^{1-Z_{kg}} \notag \\
        & \quad \times \prod_{k=1}^{K} \prod_{r=1}^{R} \alpha_k^{W_{kr}} (1-\alpha_k)^{1-W_{kr}} \times \prod_{g=1}^{G} \prod_{r=1}^{R} \beta_g^{H_{rg}} (1-\beta_g)^{1-H_{rg}} \notag \\
        & \quad \times \prod_{k=1}^{K} \alpha_k^{a_1-1}(1-\alpha_k)^{a_2-1} \notag \\
        &\quad \times\prod_{g=1}^G \left[\frac{1}{B(b_1,b_2)} \beta_g^{b_1-1}(1-\beta_g)^{b_2-1} \right]^{\psi_g}\left[\frac{1}{B(c_1,c_2)}\beta_g^{c_1-1}(1-\beta_g)^{c_2-1}\right]^{1-\psi_g}\notag\\
        & \quad \times \prod_{g=1}^G \pi^{\psi_g}(1-\pi)^{1-\psi_g} \times \pi^{d_1-1}(1-\pi)^{d_2-1}. \label{EQ:6a}
\end{align}
      \subsection{Gibbs Sampler and Maximum A Posteriori (MAP) Estimation}
    \label{subsec:inference}
    Since the full conditional distributions are available in closed form, each parameter block can be sampled directly. The sampler is summarized in Algorithm~\ref{alg:gibbs}. Note that, within a Gibbs sweep, $\pi$ is updated using the previous-iteration values of $\boldsymbol{\psi}$ before $\boldsymbol{\psi}$ is in turn updated using the new $\pi$; the order is immaterial for the validity of the sampler.

	\begin{algorithm}[htbp]
		\caption{Gibbs Sampler}
		\label{alg:gibbs}
		\begin{algorithmic}[1]
			\STATE \textbf{Input:} hyperparameters $a_1,a_2,\ b_1,b_2,\ c_1,c_2,\ d_1,d_2$
			\STATE \textbf{Initialize:} $\mathbf{W},\ \mathbf{H},\ \boldsymbol{\alpha},\ \boldsymbol{\beta},\ \boldsymbol{\psi},\ \pi$
			\FOR{iteration $= 1$ to $T$}
			\FOR{each $k$}
			\STATE Sample $\alpha_k$
			\ENDFOR
			\FOR{each $g$}
			\STATE Sample $\beta_g$
			\ENDFOR
			\FOR{each $W_{kr}$}
			\STATE Sample $W_{kr}$
			\ENDFOR
			\STATE Sample $\pi$
			\FOR{each $g$}
			\STATE Compute probabilities for $\psi_g$
			\STATE Sample $\psi_g$
			\ENDFOR
			\FOR{each $H_{rg}$}
			\STATE Sample $H_{rg}$
			\ENDFOR
			\ENDFOR
			\STATE \textbf{Return:} Posterior samples of $\,\mathbf{W},\ \mathbf{H},\ \boldsymbol{\alpha},\ \boldsymbol{\beta},\ \boldsymbol{\psi},\ \pi$
		\end{algorithmic}
	\end{algorithm} 
	%------------------------------------------------------------------------------------
    Using the Gibbs sampler of Algorithm~\ref{alg:gibbs}, we draw $T$ samples from the posterior distribution in Eq.~\eqref{EQ:6a},
    \begin{align*}
      \left\{\,\left(\mathbf{W}^{(t)},\ \mathbf{H}^{(t)},\
     \boldsymbol{\alpha}^{(t)},\ \boldsymbol{\beta}^{(t)},\ \boldsymbol{\psi}^{(t)},\
     \pi^{(t)}\right)\,\right\}_{t=1}^T.
    \end{align*}
    Because the posterior mean would yield non-binary factor matrices, we use the Maximum A Posteriori (MAP) estimate, defined as the sample attaining the largest unnormalized log-posterior. Specifically, we identify the iteration
    \begin{align*}
     t^* = \arg\max_{t=1,\ldots,T}\; \log
        P\!\left(\mathbf{W}^{(t)},\ \mathbf{H}^{(t)},\
        \boldsymbol{\alpha}^{(t)},\ \boldsymbol{\beta}^{(t)},\ \boldsymbol{\psi}^{(t)},\
        \pi^{(t)} \mid \mathbf{X}\right),
    \end{align*}
    and take the corresponding factor matrices $(\hat{\mathbf{W}}, \hat{\mathbf{H}}) = (\mathbf{W}^{(t^*)}, \mathbf{H}^{(t^*)})$ as the point estimates of the latent Boolean factors. This approach ensures that $\hat{\mathbf{W}}$ and $\hat{\mathbf{H}}$ are strictly binary, directly interpretable, and correspond to an actual posterior sample, preserving the integrity of the Boolean factorization structure. The corresponding MAP estimate of the reconstructed matrix $\mathbf{Z}$ in Eq.~\eqref{Eq:1} is $\hat{\mathbf{Z}} = \hat{\mathbf{W}} \circ \hat{\mathbf{H}}$.

    \subsection{Bipartite Graph Interpretation and Biclique Cover}

    Following \citet{miettinen2020recent}, we interpret $\mathbf{Z}\in\{0,1\}^{K\times G}$ as the bi-adjacency matrix of a bipartite graph $\mathcal{G} = (U \cup V, E)$, where $U = \{1,\dots,K\}$ and $V = \{1,\dots,G\}$. An edge $(k,g) \in E$ exists if and only if $Z_{kg} = 1$. In this representation, each rank-one Boolean component corresponds to a biclique in $\mathcal{G}$, and a Boolean matrix factorization corresponds to a (possibly overlapping) biclique cover of $\mathcal{G}$.

    \begin{proposition}[Equivalence between Boolean Matrix Factorization and Biclique Cover]
    \label{prop:bicliq_mat}
    Let $\mathbf{Z} \in \{0,1\}^{K \times G}$ and let $\mathcal{G} = (U \cup V, E)$ be the associated bipartite graph with bi-adjacency matrix $\mathbf{Z}$. Suppose $\mathbf{Z}$ admits a Boolean matrix factorization with factor matrices $\mathbf{W} \in \{0,1\}^{K \times R}$ and $\mathbf{H} \in \{0,1\}^{R \times G}$. Writing the factorization as an entry-wise OR of outer products,
    \begin{align*}
        \mathbf{W}\circ \mathbf{H} = \bigvee_{r=1}^R \mathbf{w}_r \otimes \mathbf{h}_r,
    \end{align*}
    where $\mathbf{w}_r \in \{0,1\}^{K}$ is the $r$th column of $\mathbf{W}$, $\mathbf{h}_r \in \{0,1\}^{G}$ is the $r$th row of $\mathbf{H}$, and $\mathbf{w}_r \otimes \mathbf{h}_r$ denotes their (binary) outer product. For each $r = 1, \dots, R$, define
    \begin{align*}
        U_r = \{\, k \in U : (\mathbf{w}_r)_k = 1 \,\},
        \quad
        V_r = \{\, g \in V : (\mathbf{h}_r)_g = 1 \,\}.
    \end{align*}
    Then $C_r = (U_r \cup V_r, U_r \times V_r)$ is a biclique in $\mathcal{G}$ (edges $\{u,v\}$ with $u\in U_r$, $v\in V_r$, as $\mathcal{G}$ is undirected). Moreover, the edge set satisfies
    \begin{align*}
    E = \bigcup_{r=1}^R \bigl(U_r \times V_r\bigr),
    \end{align*}
    so that $\{C_1,\dots,C_R\}$ forms a (possibly overlapping) biclique cover of $\mathcal{G}$.

    Conversely, any biclique cover $\{C_1,\dots,C_R\}$ of $\mathcal{G}$ induces a Boolean matrix factorization $\mathbf{Z} = \mathbf{W} \circ \mathbf{H}$ by setting $W_{kr} = \mathbb{1}[k \in U_r]$ and $H_{rg} = \mathbb{1}[g \in V_r]$ for all $k,r,g$.
    \end{proposition}
    %%%%%%%%%%%%%%%%%%%%%%%%%%%%%%%%%%%%%%%%%%%%%%%%%%%%%%%%%%%%%%%%%%%%%%%%%%%%%%%%%%%%%%
    \section{Evaluation Criteria and Diagnostics}
    \label{sec:diagnostics}

    In this section, we formalize the evaluation criteria used to assess the performance of the proposed Boolean Matrix Factorization method. We focus on two issues: (i) \emph{latent factor recovery}, which quantifies the accuracy with which the true underlying binary latent factors are recovered; and (ii) \emph{reconstruction equivalence and factor-recovery stability}, where reconstruction equivalence evaluates whether distinct fitted factorizations yield reconstructions that are indistinguishable at the level of the observed data, and factor-recovery stability measures the robustness and reproducibility of recovered factors across iterations (and across independent runs).

	\subsection{Latent Factor Recovery}

    Consider two Boolean factorizations of the same shape, $\mathbf{Z}' = \mathbf{W}' \circ \mathbf{H}'$ and $\mathbf{Z}'' = \mathbf{W}'' \circ \mathbf{H}''$, with $\mathbf{W}', \mathbf{W}'' \in \{0,1\}^{K \times R}$ and $\mathbf{H}', \mathbf{H}'' \in \{0,1\}^{R \times G}$. Each column $\mathbf{W}'_{\cdot r}$ and corresponding row $\mathbf{H}'_{r\cdot}$ together describe the $r$-th latent Boolean factor of $\mathbf{Z}'$, and analogously for $\mathbf{Z}''$. The latent factor recovery metric quantifies how faithfully a Boolean matrix factorization method recovers the true underlying latent structure, using three steps detailed below.

	\subsubsection*{Step 1: Similarity between individual factors}
	The pairwise similarity between rows $\mathbf{H}'_{r\cdot}$ and $\mathbf{H}''_{r'\cdot}$ of the two factor matrices is computed with the Jaccard similarity,
	\begin{align*}
		J(\mathbf{x},\mathbf{y})
		= \frac{\left\lVert \mathbf{x} \wedge \mathbf{y} \right\rVert_1}
		{\left\lVert \mathbf{x} \vee \mathbf{y} \right\rVert_1}, \qquad J(\mathbf{0},\mathbf{0}) := 1,
	\end{align*}
	where $\mathbf{x}\wedge\mathbf{y}$ and $\mathbf{x}\vee\mathbf{y}$ are elementwise \texttt{AND} and \texttt{OR}, and $\|\cdot\|_1$ is the number of ones. We align the factors based on $\mathbf{H}$ alone, since each Boolean factor is identified by its row of $\mathbf{H}$; the matching permutation is then applied to the columns of $\mathbf{W}$. The pairwise similarities are collected in
	\begin{align*}
		\mathbf{S} = (s_{rr'})_{r,r'=1}^{R} \in [0,1]^{R\times R},
		\qquad
		s_{rr'} = J(\mathbf{H}'_{r\cdot}, \mathbf{H}''_{r'\cdot}).
	\end{align*}
    
	\subsubsection*{Step 2: Factor alignment via the Hungarian algorithm}

	Since the factorization is identified only up to a permutation of the $R$ factors, we seek a permutation $\sigma$ of $\{1,\dots,R\}$ that maximizes total similarity between matched factors,
	\begin{align*}
		\max_{\sigma \in \mathcal{S}_R} \sum_{r=1}^{R} s_{r,\sigma(r)},
	\end{align*}
	where $\mathcal{S}_R$ is the set of all permutations of $\{1,\dots,R\}$. This is a standard linear assignment problem on a complete bipartite graph with edge weights $s_{rr'}$, solvable in $O(R^3)$ time by the Hungarian algorithm.

	\subsubsection*{Step 3: Aggregate factor alignment similarity (AFAS)}
    \label{subsec:AFAS}
	Let $\sigma^\star$ denote an optimal permutation. The aggregate factor alignment similarity (AFAS) is the average matched similarity,
	\begin{align*}
		\overline{s}
		= \frac{1}{R}\sum_{r=1}^{R}
		s_{r,\sigma^\star(r)}
		= \frac{1}{R}\sum_{r=1}^{R}
		J\!\big(\mathbf{H}'_{r\cdot}, \mathbf{H}''_{\sigma^\star(r)\cdot}\big).
	\end{align*}
	Because $\overline{s}$ is computed after the optimal alignment, it is invariant to any reordering of the columns of $\mathbf{W}'$ or $\mathbf{W}''$ (and the corresponding rows of $\mathbf{H}'$ or $\mathbf{H}''$); the cost is computing the full $R\times R$ similarity matrix $\mathbf{S}$ and solving an $R\times R$ assignment problem. A value of $\overline{s}$ close to $1$ indicates that each true factor is recovered as a distinct estimated factor with high fidelity; values close to $0$ indicate poor recovery.
    
    \subsection{Reconstruction Equivalence and Factor-Recovery Stability}
    Reconstruction equivalence for each MAP reconstruction $\hat{\mathbf{Z}}$ is evaluated with respect to the ground truth $\mathbf{X}_{\text{truth}}$ using the well-known metrics: specificity, F1, Matthews correlation coefficient (MCC), and reconstruction error rate. These are global goodness-of-fit measures and are secondary in simulation studies: distinct factorizations may yield almost identical reconstructions, but a good fit to $\mathbf{X}_{\text{truth}}$ does not guarantee that the true latent structure has been recovered.

    Factor-recovery stability, in contrast, is assessed by tracking the stability of the latent structure that generates the reconstruction. Using the AFAS defined in the previous subsection, for each iteration $t=1,\ldots,T$ we compare the ground-truth factor matrix $\mathbf{H}_{\text{truth}}$ with the iteration-$t$ sampled value $\mathbf{H}^{(t)}$ aligned to $\mathbf{H}_{\text{truth}}$.
	%%%%%%%%%%%%%%%%%%%%%%%%%%%%%%%%%%%%%%%%%%%%%%%%%%%%%%%%%%%%%%%%%%%%%%%%%%%
	\section{Simulation Experiments}
	\label{sec:sim_Expts}
	Using the criteria defined in Section~\ref{sec:diagnostics}, we evaluate whether the proposed \texttt{BBMF} model successfully recovers latent genomic factors from binary copy-number data, and whether realistic patterns of sparsity and prevalence in genomic alterations can be learned from a hierarchical prior rather than enforced through ad hoc constraints. Throughout this section, rows of $\mathbf{X} \in \{0,1\}^{K\times G}$ index samples ($k = 1,\dots,K$), columns index genomic loci ($g = 1,\dots,G$), and $\mathbf{X}$ arises from $R$ unobserved binary factors combining according to Boolean logic via Eq.~\eqref{Eq:1}.
    \subsection{Scenario 1: Block-Structured Genomic Factor Matrix $\mathbf{H}$}
    In this scenario, each latent factor in $\mathbf{H}$ activates contiguous chromosomal arms (in genomic order), producing broad, arm-level alteration patterns rather than sparse, scattered hits. The data are simulated as follows.
    
	\subsubsection{Latent Genomic Factors ($\mathbf{H}$)}
	The matrix $\mathbf{H} \in \{0,1\}^{R \times G}$ encodes the genomic footprint of each latent factor: the $r$-th row encodes factor $r$, and the entry $H_{rg} = 1$ indicates that factor $r$ affects locus $g$. The parameter $\theta_{rg}$ is the probability that locus $g$ is affected by factor $r$, specified as a piecewise-constant function of $g$:
    \begin{equation*}
    \theta_{rg} =
    \begin{cases}
        \theta_r^{(\mathrm{high})}, & g \in G_r, \\[4pt]
        \theta_r^{(\mathrm{low})},  & g \notin G_r,
    \end{cases}
    \qquad
    H_{rg} \mid \theta_{rg} \sim \mathrm{Bernoulli}(\theta_{rg}),
    \end{equation*}
    where $G_r \subset \{1,\dots,G\}$ is the set of chromosomal arms targeted by factor $r$, with $\theta_r^{(\mathrm{high})} \gg \theta_r^{(\mathrm{low})}$ so that large $\theta_{rg}$ yields broad factors spanning many columns and small $\theta_{rg}$ yields sparse factors affecting few columns. We set 
     \begin{align*}
     \theta_r^{(\text{high})} = (0.8,\, 0.45,\, 0.65,\, 0.05), \quad\theta_r^{(\text{low})}  = (0.05,\, 0.05,\, 0.25,\, 0.05),
     \end{align*}
     for factors $r = 1,\dots,4$, respectively. The genome is indexed by $G = 44 $chromosomal arms ordered as $1\text{p}, 1\text{q}, 2\text{p}, 2\text{q}, \dots, 22\text{p}, 22\text{q},$ so that locus \(g \in \{1,\dots,44\}\) corresponds to the \(g\)-th arm in this ordering. For each factor \(r\), the targeted set \(G_r\) is a contiguous block of arms:
    \begin{align*}
        G_1 = \{1,\dots,22\}, \qquad
        G_2 = \{20,\dots,40\}, \qquad
        G_3 = \{25,\dots,44\}, \qquad
        G_4 = \{1,\dots,44\}.
    \end{align*}
    Thus, each \(G_r\) is contiguous along the genome; the sets are not disjoint (\(G_1,G_2,G_3\) partially overlap), and \(G_4\) spans all chromosomal arms. The probabilities $\theta_r^{(\mathrm{high})}$ and $\theta_r^{(\mathrm{low})}$ control how densely each factor marks loci inside and outside its region $G_r$. For factor $r=1$, $G_1 = \{1,\dots,22\}$ with $\theta_1^{(\mathrm{high})}=0.8 \gg \theta_1^{(\mathrm{low})}=0.05$, producing a very dense block of ones in loci $1$–$22$ and almost no activity elsewhere. For factor $r=2$, $G_2 = \{20,\dots,40\}$ and $\theta_2^{(\mathrm{high})}=0.45 \gg \theta_2^{(\mathrm{low})}=0.05$ yield a moderately dense block in the overlapping mid-genome region $20$–$40$. For factor $r=3$, $G_3 = \{25,\dots,44\}$ with $\theta_3^{(\mathrm{high})}=0.65$ and a relatively large background $\theta_3^{(\mathrm{low})}=0.25$ produces a dense block in $25$–$44$ but also non-negligible activity outside this region, so the factor is less sharply localized. Finally, factor $r=4$ has $G_4 = \{1,\dots,44\}$ and $\theta_4^{(\mathrm{high})} = \theta_4^{(\mathrm{low})} = 0.05$, resulting in a sparse, globally acting factor with scattered ones across all loci (see Figure \ref{fig:Sim_Scen1_Matrices}).
    
	\subsubsection{Sample--Factor Loadings ($\mathbf{W}$)}
	The matrix $\mathbf{W} \in \{0,1\}^{K \times R}$ indicates which factors are active in which samples; $W_{kr} = 1$ means sample $k$ expresses factor $r$. We assign each factor a prevalence parameter $\eta_r$, the probability that factor $r$ is active in a randomly chosen sample, drawn as
        \begin{align*}
            \eta_r \sim \mathrm{Beta}(a^*_r, b^*_r), \qquad r = 1,\dots,R,
        \end{align*}
    and set $W_{kr} \mid \eta_r \sim \mathrm{Bernoulli}(\eta_r)$ for $k = 1,\dots,K$. The parameters $(a_r^\ast,b_r^\ast)$ control the sparsity of factor activations; in Scenario~1 we fix them to $a_r^\ast = 2$ and $b_r^\ast = 6$ for all $r = 1,\dots,R$, resulting in factors that are active in only a small fraction of samples.

    \subsubsection{Data Generation and Noise Model}
        \label{subsec:noise_model}

    Given $\mathbf{H}_{\text{truth}}$ and $\mathbf{W}_{\text{truth}}$, we first construct $\mathbf{X}_{\text{truth}} = \mathbf{W}_{\text{truth}} \circ \mathbf{H}_{\text{truth}}$. To obtain a noisy observation matrix, we apply a fixed-budget bit-flip procedure with noise level $p_{\text{flip}} = 0.20$. Specifically, letting $N = K G$ denote the total number of entries in $\mathbf{X}_{\text{truth}}$, we compute $N_{\text{flip}} = \lceil p_{\text{flip}} N \rceil$ and select $N_{\text{flip}}$ distinct indices uniformly at random (without replacement). For each selected index $(k,g)$ we set $X_{\text{sim},kg} = 1 - X_{\text{truth},kg}$, while all remaining entries are left unchanged; this yields the simulated data matrix $\mathbf{X}_{\text{sim}}$. Marginally, each entry is flipped with probability approximately $p_{\text{flip}}$, so the noise distribution is close to the \texttt{BBMF} likelihood in Section~\ref{sec:proposed_BBMF} with $p_{11} \approx 0.80$ and $p_{10} \approx 0.20$, but the total number of flipped entries is fixed rather than random.

\begin{figure}[tbp]
		\centering
		\resizebox{\textwidth}{!}{%
			\begin{tabular}{ccccccc}  % 7 columns
				\adjustbox{valign=m}{\includegraphics[height=5cm,width=.85cm]{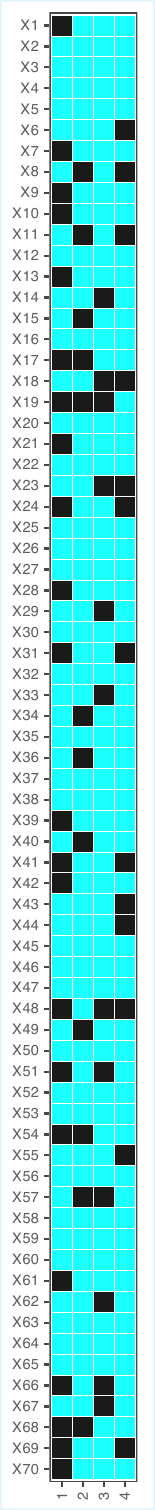}} &
				\adjustbox{valign=m}{$\circ$} &
				\adjustbox{valign=m}{\includegraphics[height=.85cm,width=3.5cm]{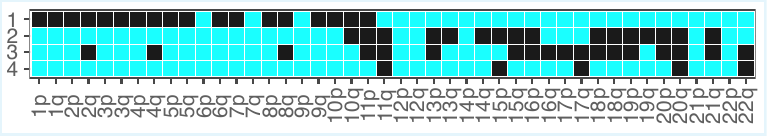}} &
				\adjustbox{valign=m}{$=$} &
				\adjustbox{valign=m}{\includegraphics[height=5cm,width=3.5cm]{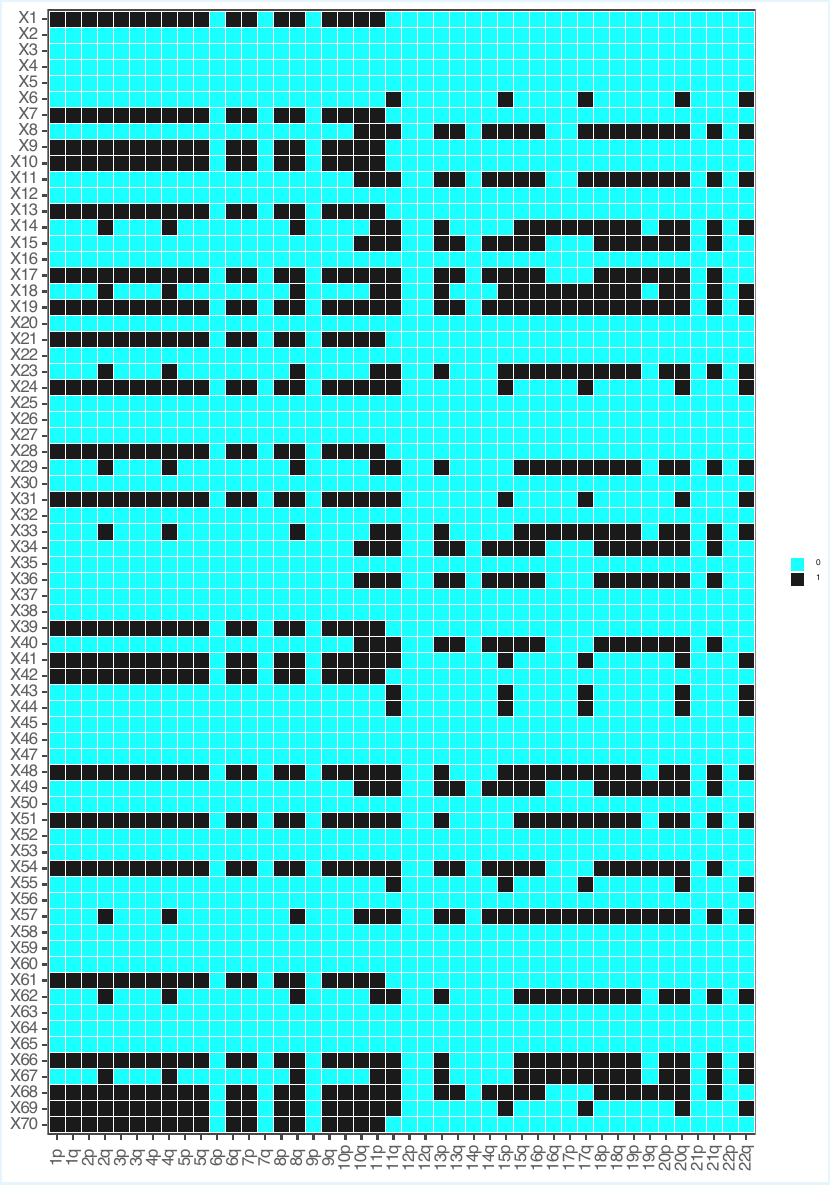}} &
				\adjustbox{valign=m}{$\xrightarrow{\substack{20\% \; \text{noise}}}$} &
				\adjustbox{valign=m}{\includegraphics[height=5cm,width=3.5cm]{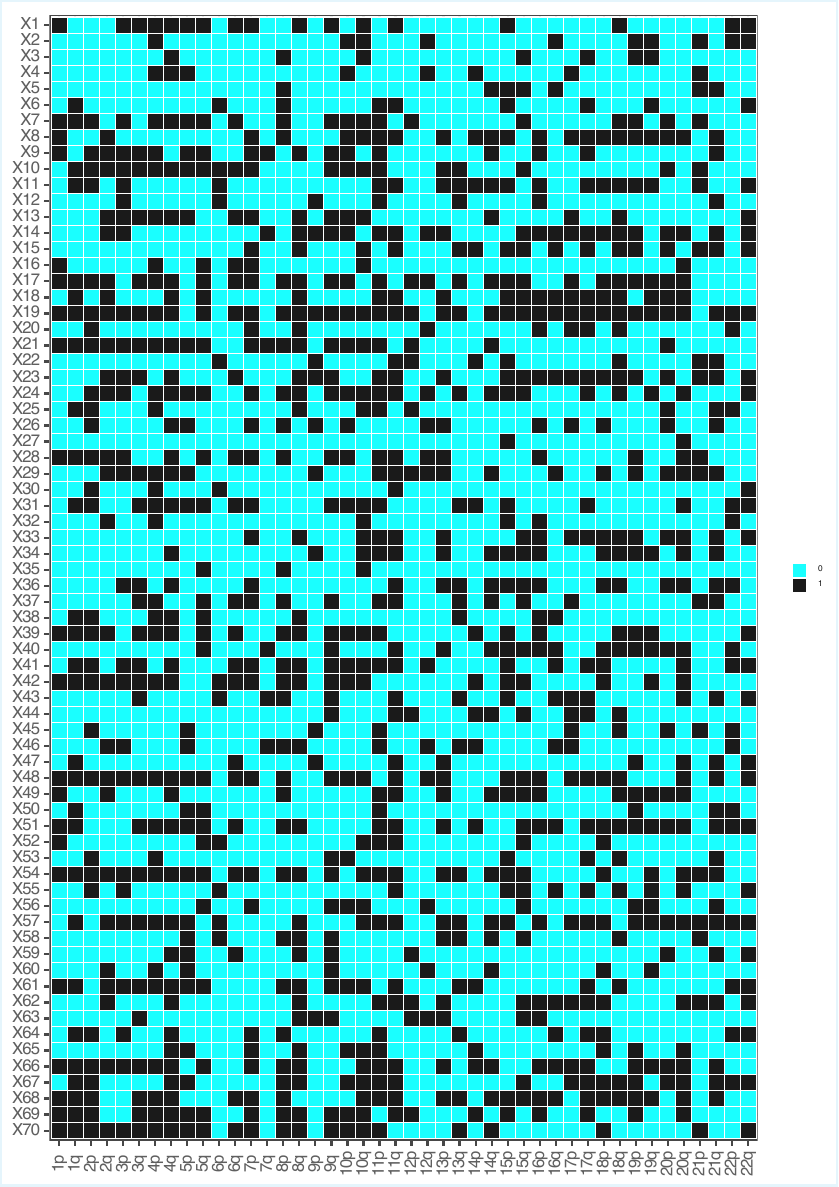}} \\
				$\mathbf{W}_{\text{truth}}$ && $\mathbf{H}_{\text{truth}}$ && $\mathbf{X}_{\text{truth}}$ &
				& $\mathbf{X}_{\text{sim}}$
			\end{tabular}%
		}
		\caption{Scenario 1: simulation setup used to generate $\mathbf{X}_{\text{truth}}$ and $\mathbf{X}_{\text{sim}}$. 
			Entries are binary, with black $=1$ (activation) and cyan $=0$ (no activation). $\mathbf{W}_{\text{truth}}$ displays a sparse, mixed activation pattern across patients, $\mathbf{H}_{\text{truth}}$ encodes contiguous block-structured genomic factors, and $\mathbf{X}_{\text{truth}}$ reflects the broad, arm-level alteration patterns induced by the Boolean product. The noise in $\mathbf{X}_{\text{sim}}$ partially obscures these blocks while leaving the dominant structure discernible.
			}
		\label{fig:Sim_Scen1_Matrices}
	\end{figure}

    For Scenario~1 we use $K = 70$ patients, $G = 44$ chromosomal arms, and $R = 4$ latent factors. Scenario~1 is summarized in Figure~\ref{fig:Sim_Scen1_Matrices}.

    %--------------------------------------------------------------------------------------

    \subsection{Scenario 2: Synthetic Data}
    \label{subsec:synthetic_data}
\begin{figure}[htbp]
		\centering
		\resizebox{\textwidth}{!}{%
			\begin{tabular}{ccccccc}
				\adjustbox{valign=m}{\includegraphics[height=4.5cm,width=.8cm]{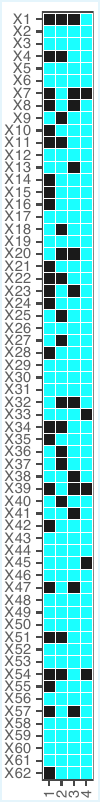}} &
				\adjustbox{valign=m}{$\circ$}&
				\adjustbox{valign=m}{\includegraphics[height=.85cm,width=3.5cm]{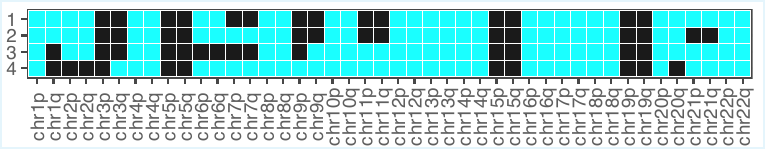}}&
				\adjustbox{valign=m}{$=$}&
				\adjustbox{valign=m}{\includegraphics[height=4.5cm,width=3.5cm]{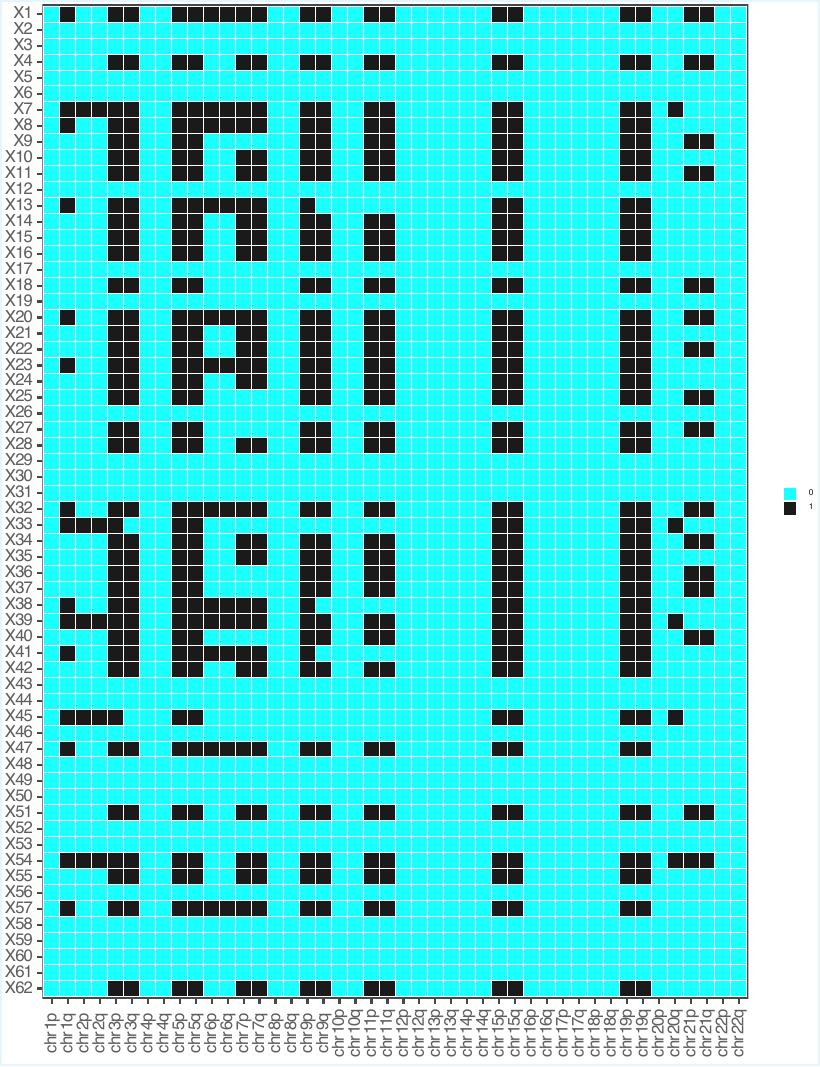}} &
				\adjustbox{valign=m}{$\xrightarrow{\substack{20\% \; \text{noise}}}$}&
				\adjustbox{valign=m}{\includegraphics[height=4.5cm,width=3.5cm]{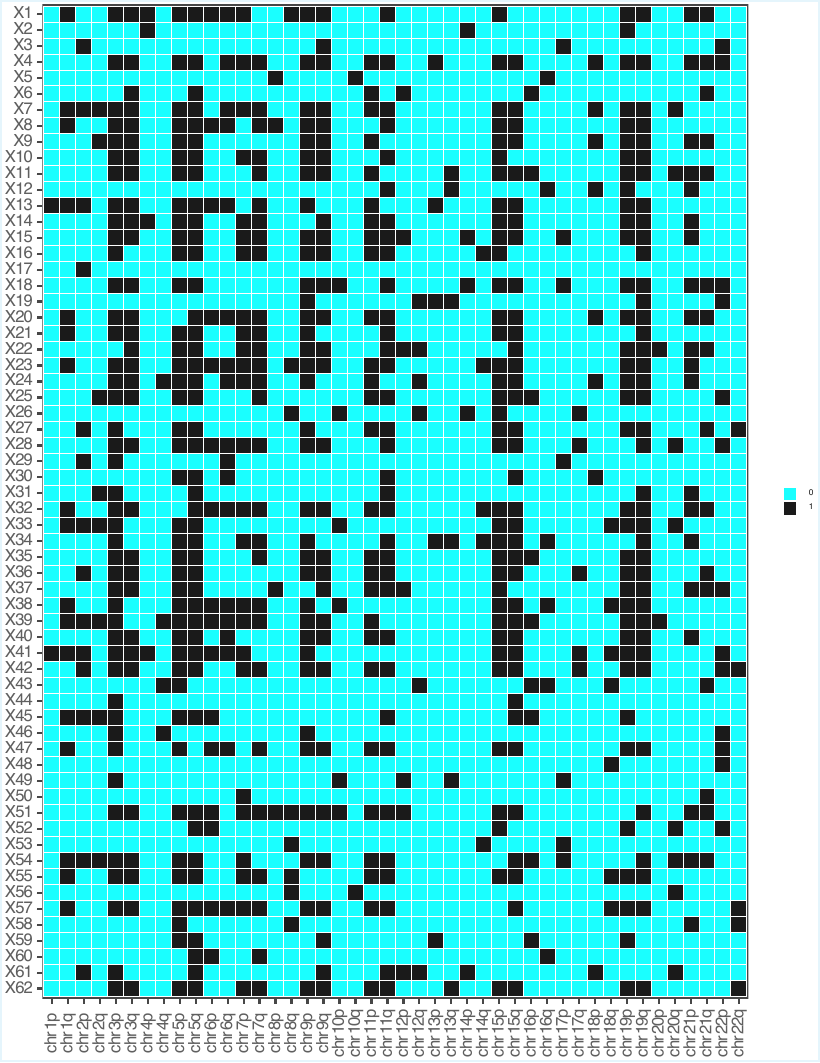}} \\
				$\mathbf{W}_{\text{truth}}$   & & $\mathbf{H}_{\text{truth}}$&& $\mathbf{X}_{\text{truth}}$& &$\mathbf{X}_{\text{sim}}$
			\end{tabular}
		}
		\caption{Scenario 2: generation of synthetic data $\mathbf{X}_{\text{truth}}$ and $\mathbf{X}_{\text{sim}}$. In $\mathbf{W}_{\text{truth}}$ and $\mathbf{H}_{\text{truth}}$, distinct vertical and horizontal blocks correspond to a few latent factors with localized support, while $\mathbf{X}_{\text{truth}}$ shows large, regular rectangles of uniform intensity, reflecting coherent factor-driven structure. The $\mathbf{X}_{\text{sim}}$ contains partially visible rectangles disrupted by many irregular entries; random perturbations obscure but do not fully erase the underlying signal.}
		\label{fig:Sim_Synth_Expt1_Matrices}
	\end{figure}

    Scenario 2 uses a synthetic binary matrix designed to mimic the properties of copy-number alteration (CNA) profiles in multiple myeloma. We first fit the Bayesian binary factor model in Eq.(\ref{EQ:6a}) to the real MM CNA data using a Gibbs sampler. After discarding burn-in, we randomly select one posterior draw $(\mathbf{W}, \mathbf{H})$ from the Markov chain and treat it as the ``true'' factorization $(\mathbf{W}_{\text{truth}}, \mathbf{H}_{\text{truth}})$. This yields factor matrices $\mathbf{W}_{\text{truth}} \in \{0,1\}^{K \times R}$ and $\mathbf{H}_{\text{truth}} \in \{0,1\}^{R \times G}$ with $K = 62$ patients, $G = 44$ chromosomal arms, and $R = 4$ latent factors. Following the procedure of Section~\ref{subsec:noise_model}, we construct $\mathbf{X}_{\text{truth}} = \mathbf{W}_{\text{truth}} \circ  \mathbf{H}_{\text{truth}}$ and then add bit-flip noise to obtain the simulated data matrix $\mathbf{X}_{\text{sim}} \in \{0,1\}^{62 \times 44}$. Scenario 2 is summarized in Figure~\ref{fig:Sim_Synth_Expt1_Matrices}.

	\subsection{Application of BBMF to the Simulated Data}
	\label{subsec:single_fact_sim_data}

    We apply the proposed \texttt{BBMF} model, the \texttt{Asso} algorithm \cite{Asso_4479462}, and \texttt{GreConD+} \cite{Belohlavek2010jcss} to each simulated data set $\mathbf{X}_{\text{sim}}$ and evaluate performance using the criteria in  Section~\ref{sec:diagnostics}. Although a comparison with the \texttt{OrMachine} \citep{Rukat2017pmlr} would have been informative, its \texttt{Python} package is no longer maintained and could not be included in the comparison.
    
    For \texttt{BBMF} we initialized the factor matrices $(\mathbf{W},\mathbf{H})$ at the solution returned by \texttt{Asso}. This provides a sensible starting point and allows us to assess whether the MCMC explores factorisations that improve on the heuristic \texttt{Asso} fit. We used weakly informative Beta priors for all model hyperparameters, setting $(a_1,a_2) = (1,1)$, $(b_1,b_2) = (1,1)$, $(c_1,c_2) = (1,1)$, and $(d_1,d_2) = (1,1)$ as in Section~\ref{sec:proposed_BBMF}. For each scenario we ran $n_{\text{chains}} = 4$ parallel MCMC chains with $n_{\text{iter}} = 100,000$ iterations per chain, and discarded the first $20,000$ iterations as burn-in after assessing convergence.
	
	\subsubsection{Latent Factor Recovery for the Simulated Data}
\begin{figure}[htbp]
    \centering
    %--- Subfigure (a) ---
    \begin{subfigure}{\textwidth}
        \centering
        \includegraphics[width=\textwidth]{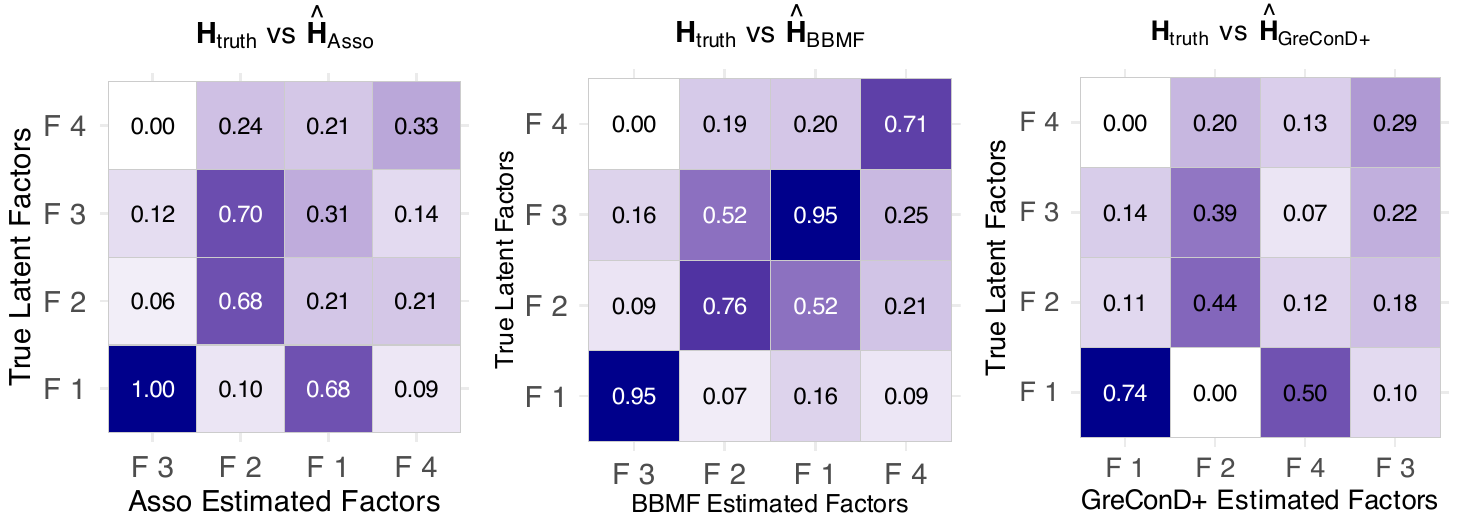}
        \caption{Scenario 1: similarity heatmaps.}
        \label{fig:similarity_H_ExptI}
    \end{subfigure}

    \vspace{0.5em}

    %--- Subfigure (b) ---
    \begin{subfigure}{\textwidth}
        \centering
        \includegraphics[width=\textwidth]{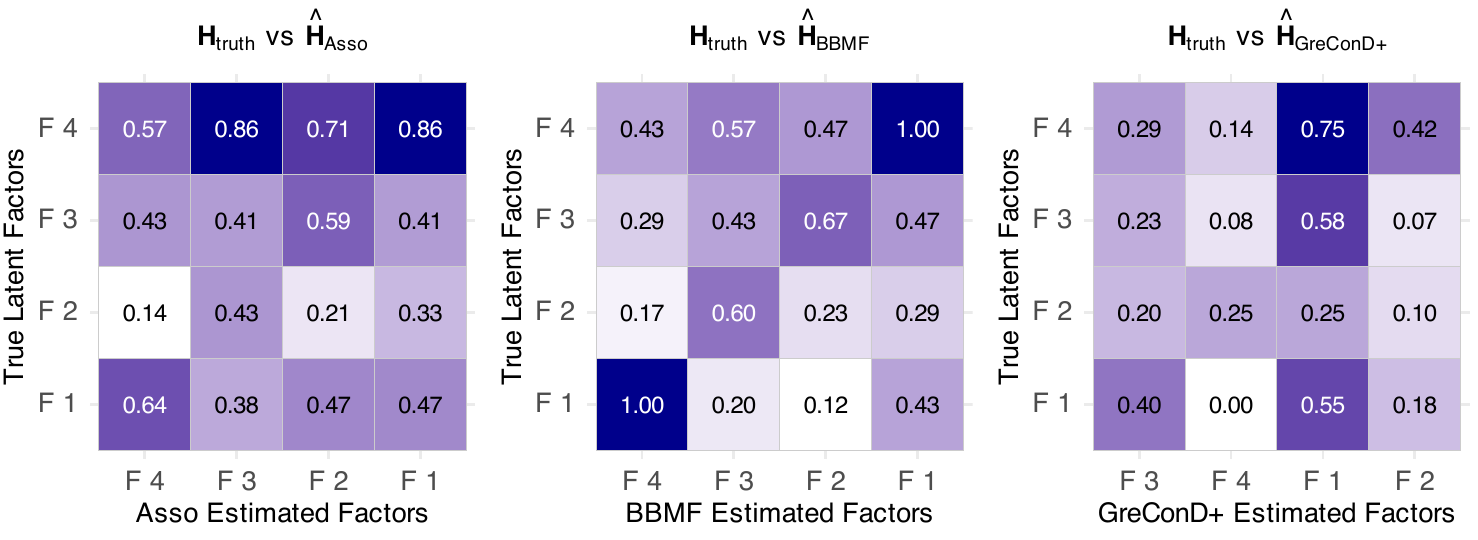}
        \caption{Scenario 2: similarity heatmaps.}
        \label{fig:sim_Synth_H1}
    \end{subfigure}

    \caption{Comparison of similarity structures between true latent factors and those recovered by \texttt{Asso}, \texttt{GreConD+}, and \texttt{BBMF}. Factors are labelled F1, F2, F3, and F4.}
    \label{fig:combined_similarity}
\end{figure}

	To assess the quality of factor recovery, we compared the estimated factor matrices from \texttt{BBMF} ($\hat{\mathbf{H}}_{\text{BBMF}}$), \texttt{Asso} ($\hat{\mathbf{H}}_{\text{Asso}}$), and \texttt{GreConD+} ($\hat{\mathbf{H}}_{\text{GreConD+}}$) with the corresponding ground-truth matrix $\mathbf{H}_{\text{truth}}$. For each method, we computed the pairwise Jaccard similarity between the estimated factor matrix and the ground truth and then visually inspected the aligned factor matrices. Alignment was performed via the Hungarian algorithm implemented in the \texttt{R} package \texttt{clue} by \citet{hornik2005clue}. The results are shown in Figure~\ref{fig:combined_similarity}.

    Figures~\ref{fig:similarity_H_ExptI} and~\ref{fig:sim_Synth_H1} display heatmaps of the Jaccard similarity metrics between the ground-truth latent factors and those estimated by each method, where each panel compares the true factors (rows) to the recovered factors (columns). \texttt{BBMF} yields the most accurate recovery, characterized by a strongly diagonal similarity matrix: the main diagonal exhibits values close to $1$ (including a perfect correspondence for factor F2), while off-diagonal entries are negligible. This pattern indicates that each true factor is associated with a single, well-identified estimated factor, with minimal cross-loading or confusion between components. In contrast, \texttt{Asso} recovers the true factors only at a coarse level, showing moderate diagonal entries accompanied by substantial off-diagonal similarities across multiple estimated components; this behavior suggests that some true factors are only partially retrieved or are conflated with others. \texttt{GreConD+} captures the dominant latent structure but frequently fragments individual true factors across several estimated components, resulting in redundant, rather than distinct and well-aligned, representations.

\begin{figure}[htbp]
		\centering
            %--- Subfigure (a) ---
    \begin{subfigure}{\textwidth}
        \centering
		\includegraphics[width=\textwidth]{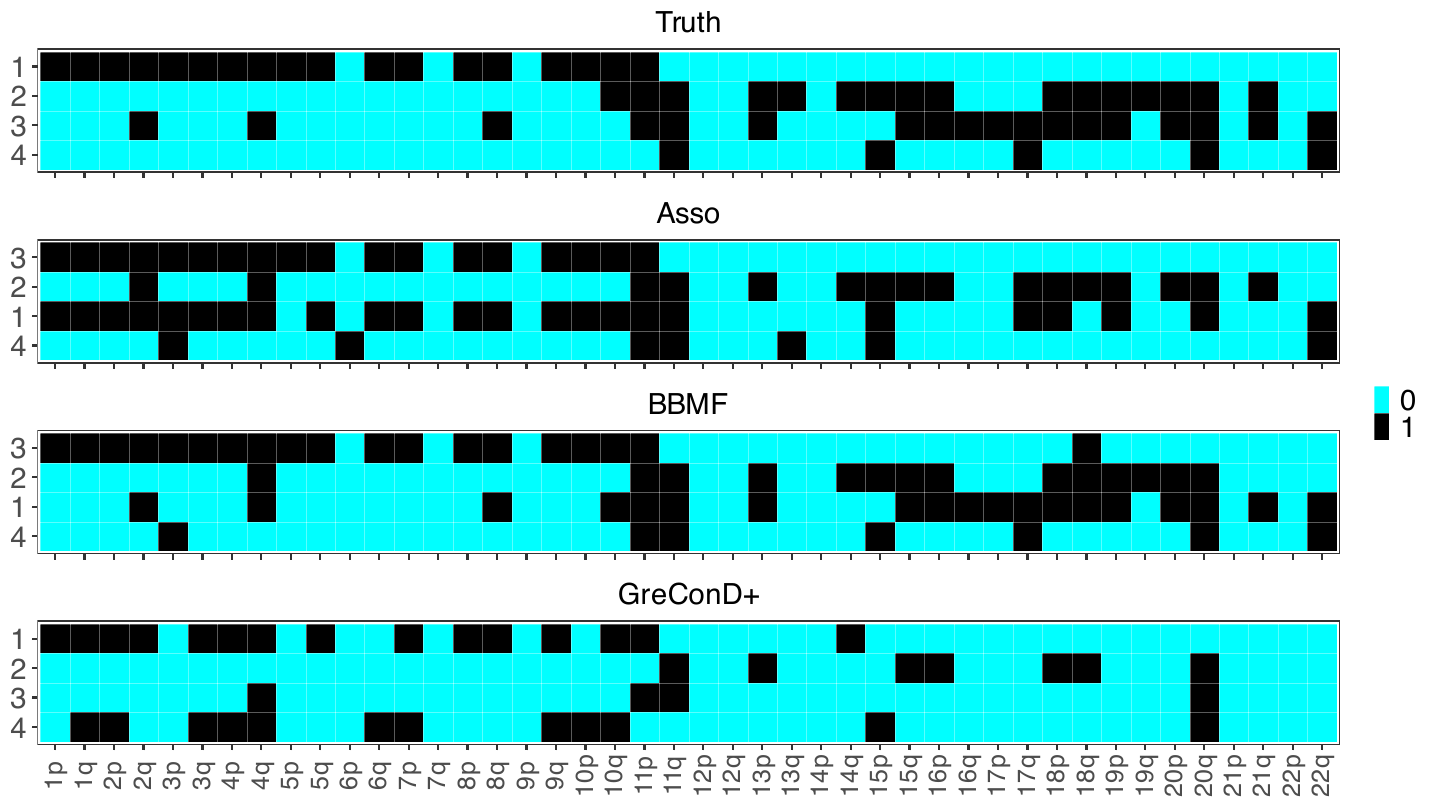}
		\caption{Scenario 1.}
		\label{fig:Asso_H_vs_Expt1_BBMF}
	 \end{subfigure}
     %--- Subfigure (b) ---
    \begin{subfigure}{\textwidth}
		\centering
		\includegraphics[width=\textwidth]{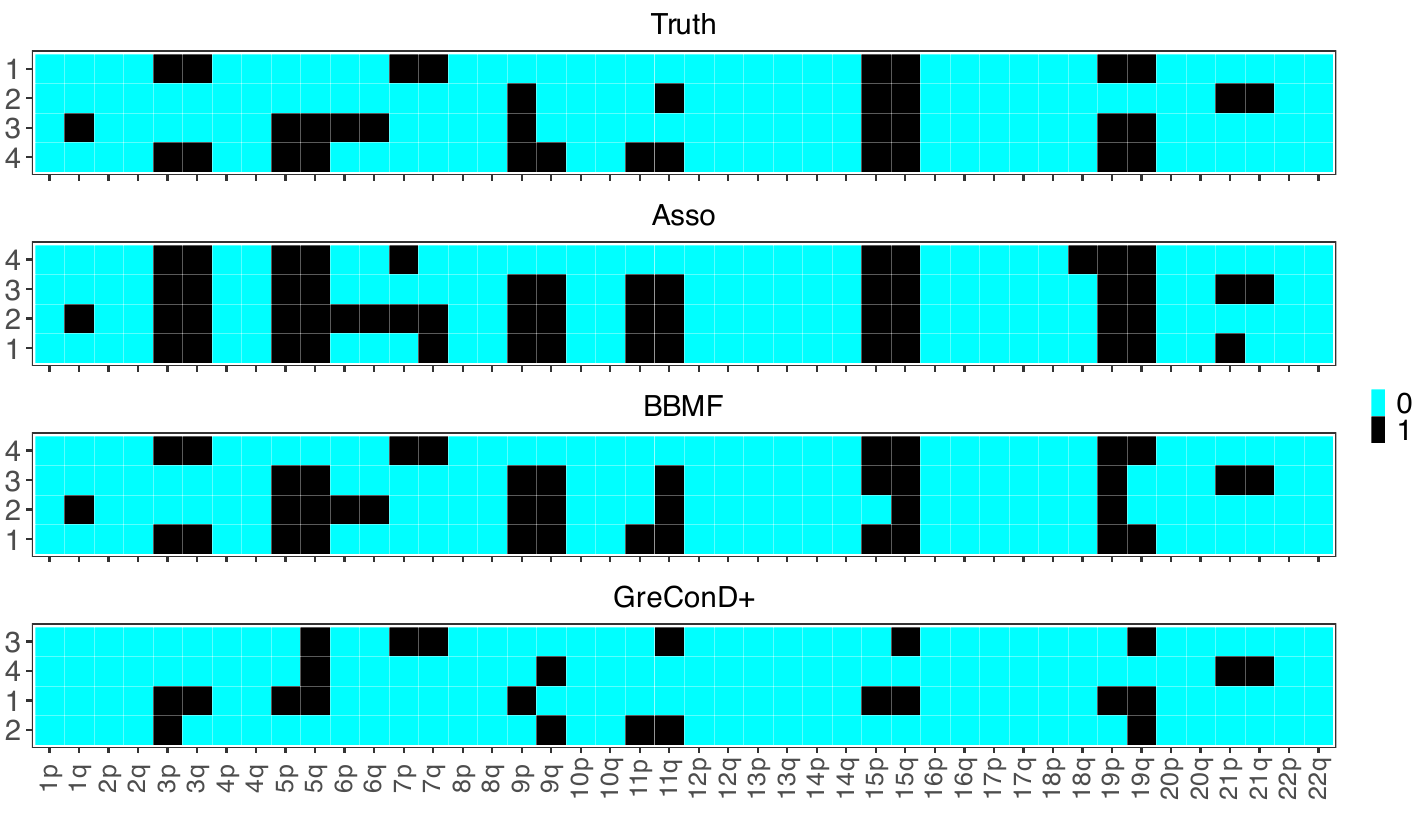}
		\caption{Scenario 2.}
		\label{fig:OrderedHs_SynthDataX_BBMF}
        	 \end{subfigure}
         \caption{Comparison of true and recovered binary factors for the two simulated scenarios.
			Within each panel, the rows representing the factors have been reordered to optimise agreement with the truth, and the ordered true factors are shown alongside the correspondingly ordered factors recovered by \texttt{Asso}, \texttt{BBMF} and \texttt{GreConD+}.}
		\label{fig:Ordered_Hs_all}    
	\end{figure}

	Next, we visualize the corresponding factor matrices after applying the permutations that best align the estimated components with the ground truth. The visualizations are presented in Figure~\ref{fig:Ordered_Hs_all}. 
    
    In Figure~\ref{fig:Asso_H_vs_Expt1_BBMF} (Scenario 1), \texttt{Asso} and \texttt{GreConD+}'s estimated factors contain many spurious activations and omissions, with columns active in the ground truth appearing only partially or inconsistently, reflecting considerable mixing between components. \texttt{BBMF}, by contrast, recovers the true binary patterns much more faithfully, with activation blocks that align closely with the ground truth and relatively few spurious entries. In Figure~\ref{fig:OrderedHs_SynthDataX_BBMF} (Scenario 2), both \texttt{Asso} and \texttt{BBMF} reproduce the main activation blocks with the correct row- and column-specific structure, with the reordering bringing both methods into close agreement with the ground truth. However, \texttt{GreConD+} produced the poorest recovery compared to the two other methods.

	\subsubsection{Reliability of the Factor Recovery by BBMF}
\begin{figure}[htbp]
        \centering
    % Row 1
        \begin{subfigure}[t]{0.48\textwidth}
            \centering
            \includegraphics[width=6cm, height=5cm]{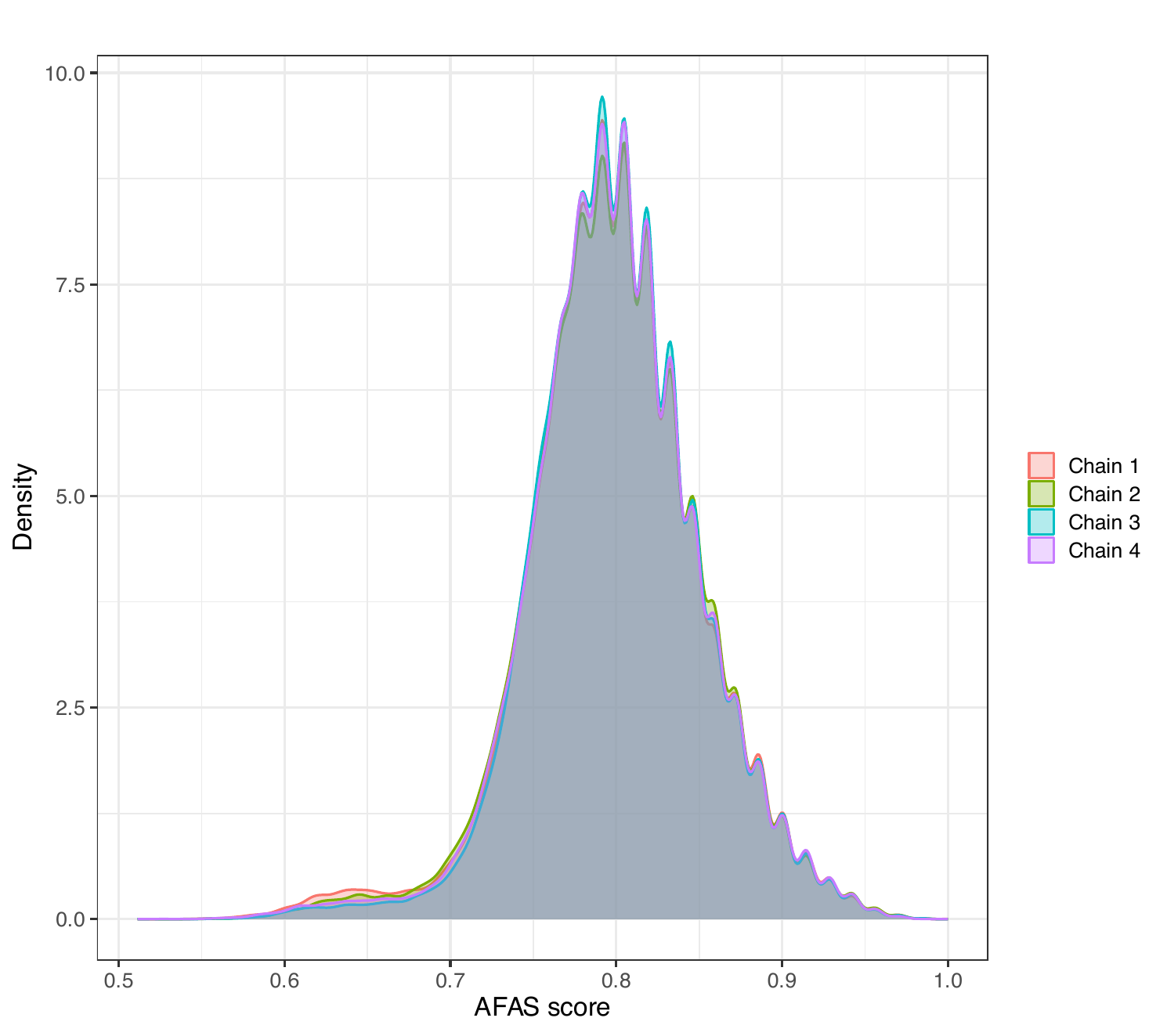}
            \caption{Scenario 1: Density of the Aggregate Factor Alignment Similarity for each chain.}
            \label{fig:post_Mean_Sim_Expt1}
         \end{subfigure}
    \hfill
        \begin{subfigure}[t]{0.48\textwidth}
            \centering
            \includegraphics[width=6cm, height=5cm]{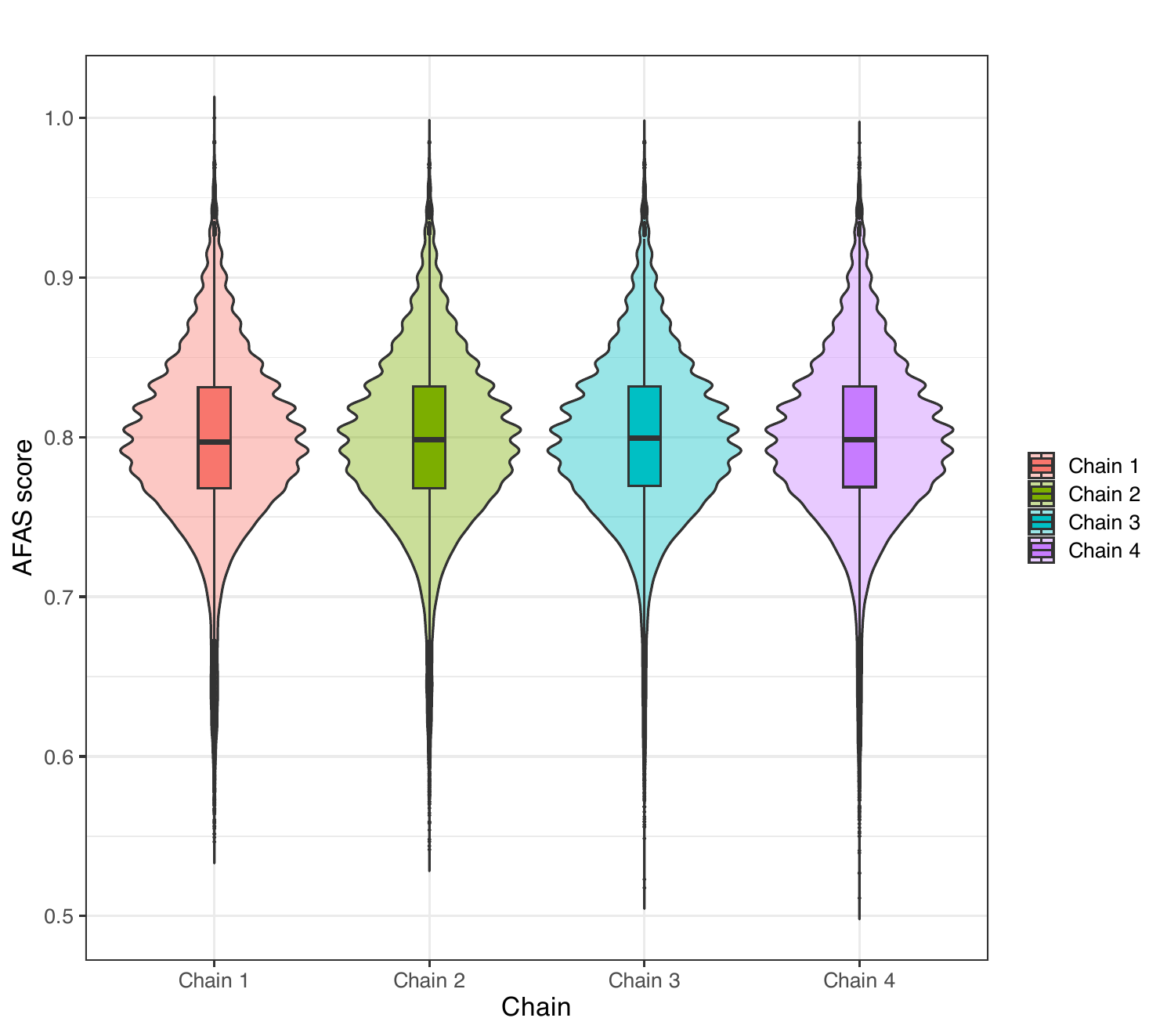}
            \caption{Scenario 1: Violin and boxplots of the Aggregate Factor Alignment Similarity by chain.}
            \label{fig:Violin_Mean_Expt1}
        \end{subfigure}
    \vspace{0.4cm}
    
    % Row 2
         \begin{subfigure}[t]{0.48\textwidth}
            \centering
             \includegraphics[width=6cm, height=5cm]{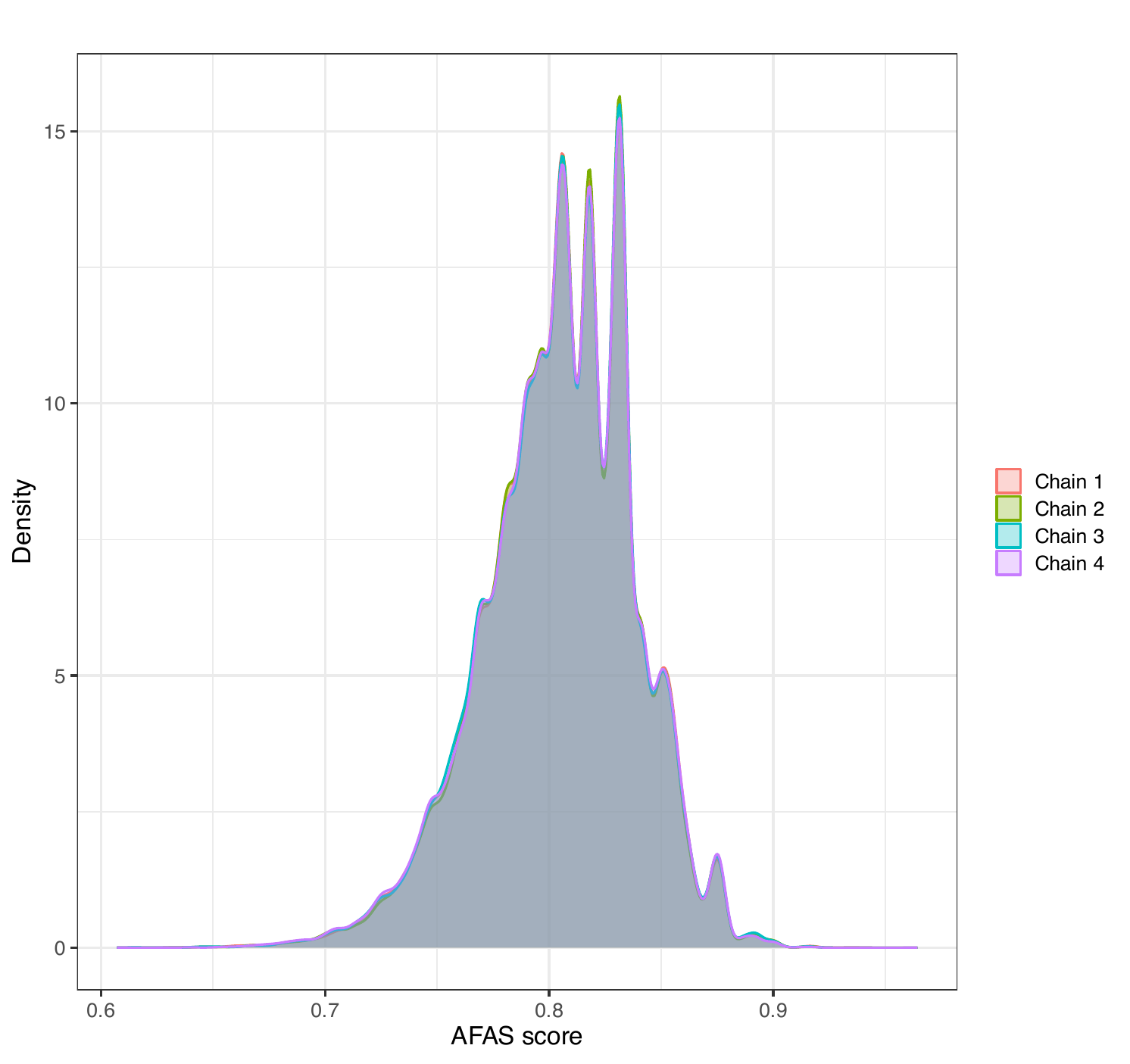}
             \caption{Scenario 2: Density of the Aggregate Factor Alignment Similarity for each chain.}
                \label{fig:Post_dist_sim_Synth}
            \end{subfigure}
    \hfill
              \begin{subfigure}[t]{0.48\textwidth}
                \centering
                 \includegraphics[width=6cm, height=5cm]{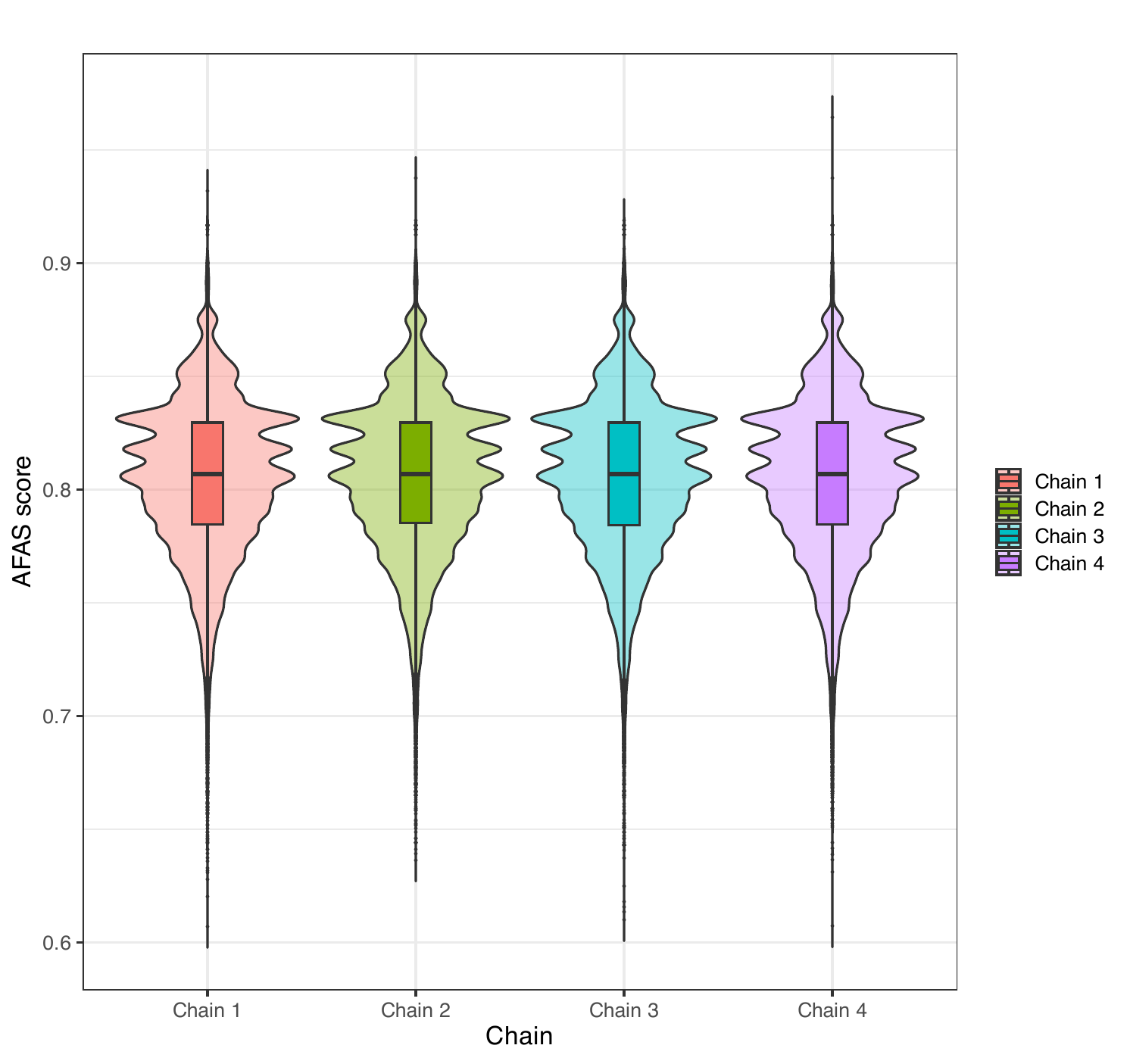}
                \caption{Scenario 2: Violin and boxplots of the Aggregate Factor Alignment Similarity by chain.}
                \label{fig:Violin_Synth_Data}
            \end{subfigure}
    
    \caption{Chain-level similarity diagnostics for factor recovery: the AFAS between $\mathbf{H}^{(t)}$ and $\mathbf{H}_{\text{truth}}$ across MCMC iterations, for the BBMF runs on the Scenario 1 and Scenario 2 simulated data.}
    \label{fig:violin_density_all_data}
    \end{figure}

	To evaluate the reliability with which \texttt{BBMF} recovers the underlying latent factors, we monitored the aggregate factor alignment similarity (AFAS; see Subsection~\ref{subsec:AFAS}) between the iteration-$t$ sampled value $\mathbf{H}^{(t)}$ and the ground-truth factor matrix $\mathbf{H}_{\text{truth}}$ at each iteration of the sampler. At each iteration, the recovered factors were first aligned to $\mathbf{H}_{\text{truth}}$, and the AFAS was then computed by averaging the matched-pair Jaccard similarities. We assess similarity diagnostics for factor recovery by (i) examining the chain's empirical distribution of AFAS values and (ii) comparing chains using violin plots with overlaid boxplots. These diagnostics are shown in Figure~\ref{fig:violin_density_all_data}.

	As shown in Figures~\ref{fig:post_Mean_Sim_Expt1} and~\ref{fig:Violin_Mean_Expt1}, the empirical densities and violin/boxplots of the AFAS for Scenario 1 largely overlap across chains and are centred around similar values, with comparable medians and interquartile ranges. These diagnostics indicate that the AFAS summary statistic agrees across chains; we caution that this single summary does not by itself rule out multimodality in the underlying parameter space.

    Similarly, Figures~\ref{fig:Post_dist_sim_Synth} and~\ref{fig:Violin_Synth_Data} for Scenario 2 show that all chains concentrate most of their AFAS mass on high values, with densities largely overlapping across chains; minor differences in spread are attributable to Monte Carlo variability.
	\subsubsection{Matrix Reconstruction Accuracy}
    Reconstruction accuracy is assessed by comparing the estimated matrices $\hat{\mathbf{Z}}_{\text{BBMF}}$, $\hat{\mathbf{Z}}_{\text{Asso}}$, and $\hat{\mathbf{Z}}_{\text{GreConD+}}$ against the ground-truth matrix $\mathbf{X}_{\text{truth}}$ using specificity ($\mathrm{TN}/(\mathrm{TN}+\mathrm{FP})$), F1 ($2\,\mathrm{precision}\cdot\mathrm{recall}/(\mathrm{precision}+\mathrm{recall})$), Matthews correlation coefficient (MCC), and the reconstruction error rate $(\mathrm{FP}+\mathrm{FN})/(KG)$. The results for both simulation scenarios are reported in Table~\ref{tab:combined_metrics}.

    \begin{table}[htbp]
        \centering
        \caption{Performance comparison of Boolean matrix factorization methods}
        \vspace{.25cm}
        \label{tab:combined_metrics}
        \begin{tabular}{llcccc}
        \hline
        Scenario & Method & Specificity & F1 & MCC & Recon.\ Error Rate \\
        \hline
        \multirow{3}{*}{Scenario 1} 
                  & Asso      & 0.910 & 0.830 & 0.767 & 0.095 \\
                  & BBMF      & 0.960 & 0.928 & 0.903 & 0.039 \\
                  & GreConD+  & 0.995 & 0.546 & 0.542 & 0.166 \\
        \hline
        \multirow{3}{*}{Scenario 2} 

                  &Asso       & 0.968 & 0.910 & 0.887 & 0.037\\
                  &BBMF       & 0.984 & 0.952 & 0.940 & 0.019\\
                  &GreConD+   & 0.999 & 0.857 & 0.839 & 0.050\\
                  \hline
        \end{tabular}
    \end{table}

    The results in Table~\ref{tab:combined_metrics} show that \texttt{BBMF} achieves the highest F1 and MCC values and the lowest reconstruction error rate in both scenarios, while \texttt{Asso} performs reasonably well across all metrics. \texttt{GreConD+} attains the highest specificity, but at the cost of substantially lower F1 and MCC scores and the highest reconstruction error rate, suggesting that it is overly conservative in predicting positive entries. In Scenario~2, all three methods improve and the relative ranking remains unchanged, with \texttt{BBMF} maintaining its advantage in all metrics and \texttt{GreConD+} continuing to show the same conservative behavior.
    
\begin{figure}[tbp]
		\centering
        %--- Subfigure (a) ---
    \begin{subfigure}{\textwidth}
        \centering
		\includegraphics[width=\textwidth]{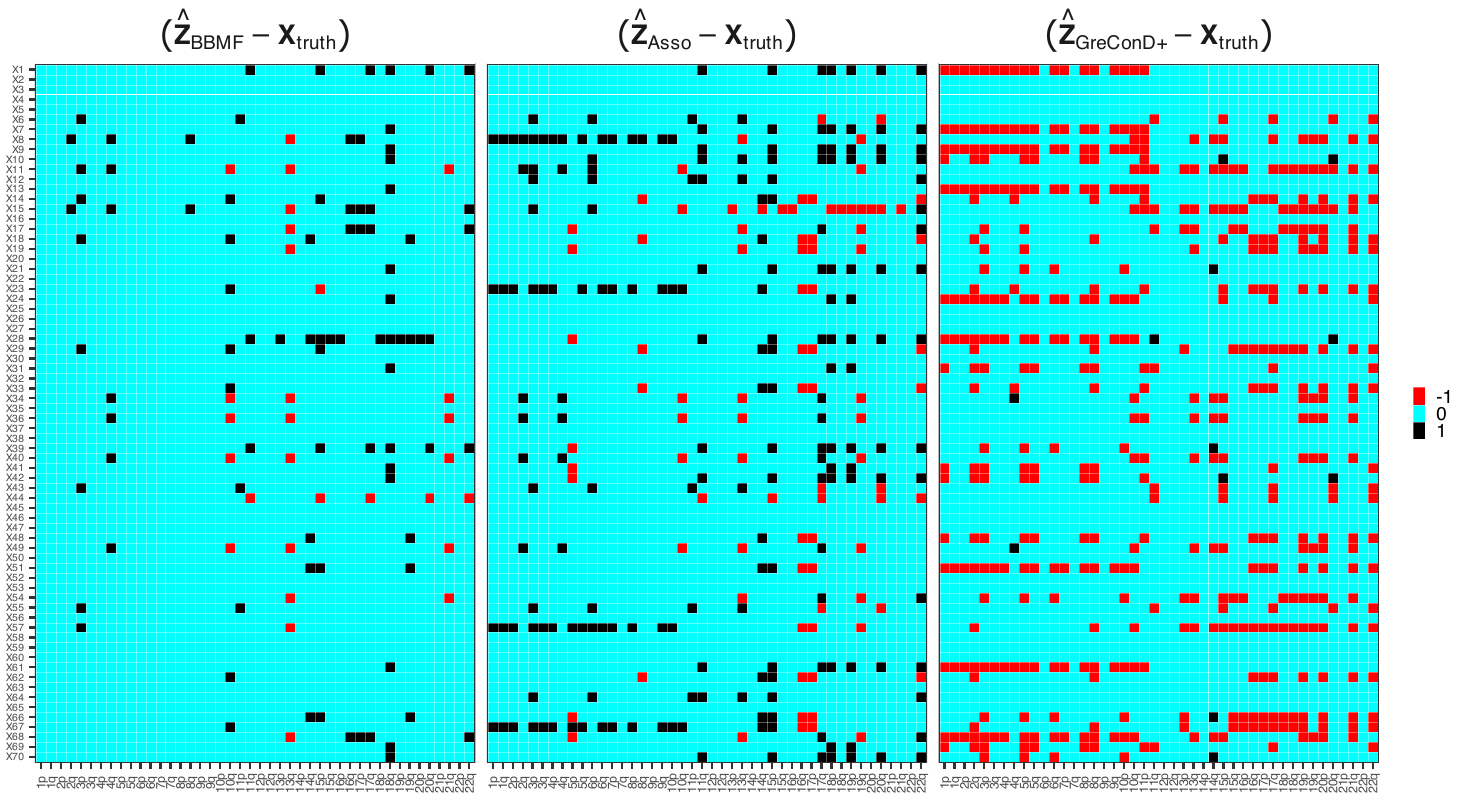}%
        \caption{Scenario 1.}
		\label{fig:Recon_X_diff_Expt1}
	\end{subfigure}

     \vspace{0.5em}

    %--- Subfigure (b) ---
    \begin{subfigure}{\textwidth}
		\includegraphics [width=\textwidth]{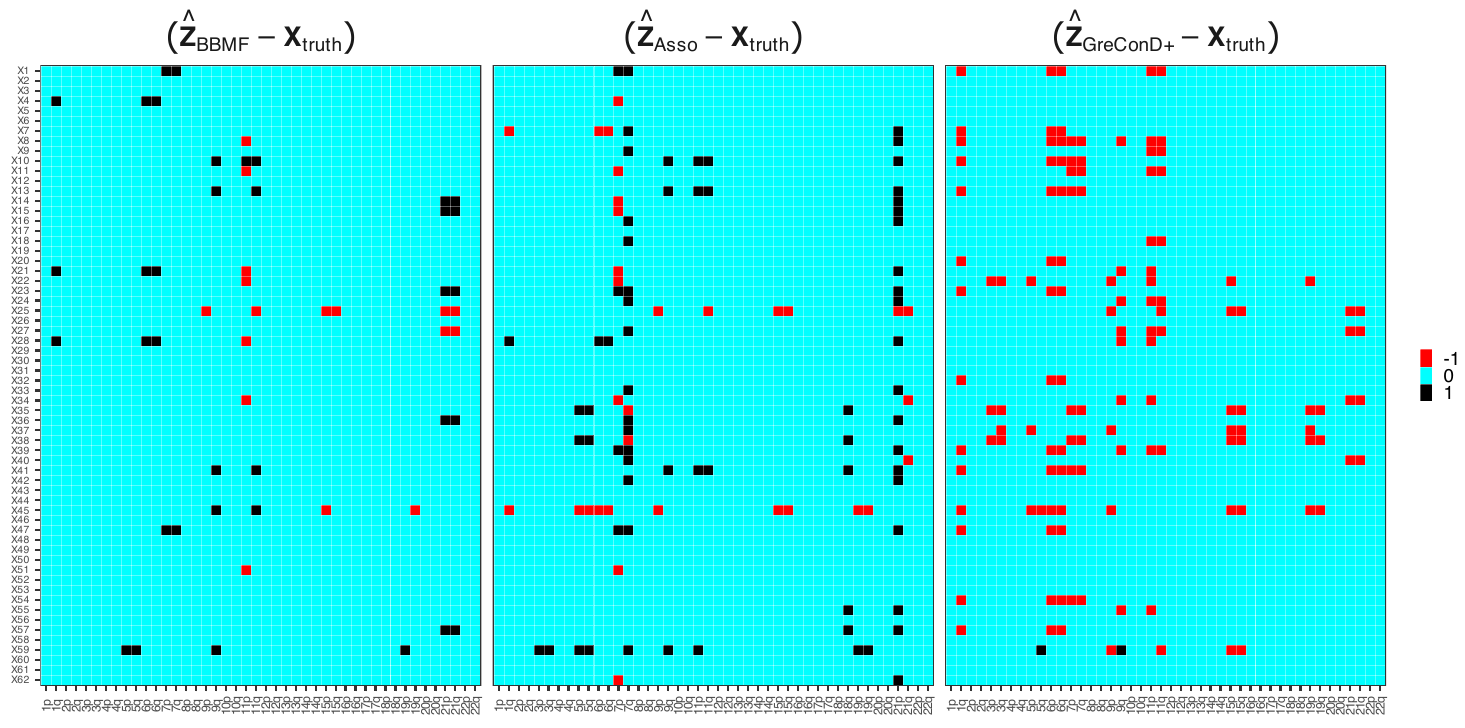}
		\caption{Scenario 2.}
		\label{fig:X_diff_synth}
	\end{subfigure}
    \caption{Heatmaps of entrywise differences between reconstructed and ground‑truth Boolean matrices for the simulation experiment. The cyan cells denote exact agreement $(0)$, black cells overestimation $(+1)$, and red cells underestimation $(-1)$.}
		\label{fig:X_diffs_all}
    \end{figure}

    Beyond the aggregate metrics, the element-wise discrepancies between the ground-truth matrices and their corresponding reconstructions from \texttt{BBMF} and \texttt{Asso} reveal the spatial organization of errors, whether mismatches are sporadic or exhibit systematic patterns across patients or chromosomal arms. The resulting heatmaps are shown in Figure~\ref{fig:X_diffs_all}.

    The elementwise difference maps $(\hat{\mathbf{Z}} - \mathbf{X}_{\text{truth}})$ shown in Figures \ref{fig:Recon_X_diff_Expt1} and \ref{fig:X_diff_synth} for \texttt{BBMF} and \texttt{Asso} in both scenarios indicate that \texttt{BBMF} systematically yields reconstructions that more accurately preserve the block structure and sparsity pattern of $\mathbf{X}_{\text{truth}}$. Specifically, the \texttt{BBMF}-based reconstructions exhibit only a small number of isolated false positives and false negatives scattered throughout the matrix. In contrast, the reconstructions obtained with \texttt{Asso} display a substantially higher density of discrepancies, including dispersed errors within multiple blocks as well as localized clusters of both spurious and missing ones in particular rows and columns.
    %-----------------------------------------------------------------------------------------
	Table~\ref{tab:diff_freq_combined} quantifies these visual impressions by presenting the frequencies and percentages of the individual differences.
    
 \begin{table}[htbp]
    \centering
    \caption{Frequency and percentage of elementwise differences 
    $(\hat{\mathbf{Z}} - \mathbf{X}_{\text{truth}})$ for \texttt{BBMF}, 
    \texttt{Asso}, and \texttt{GreConD+} methods.}
    \vspace{.25cm}
    \label{tab:diff_freq_combined}
    %\begin{tabular}{lcrrrrrr}
    \begin{tabular}{lccccccc}
        \toprule
        & & \multicolumn{2}{c}{$(\hat{\mathbf{Z}}_{\text{BBMF}} - \mathbf{X}_{\text{truth}})$}
          & \multicolumn{2}{c}{$(\hat{\mathbf{Z}}_{\text{Asso}} - \mathbf{X}_{\text{truth}})$}
          & \multicolumn{2}{c}{$(\hat{\mathbf{Z}}_{\text{GreConD+}} - \mathbf{X}_{\text{truth}})$} \\
        \cmidrule(lr){3-4} \cmidrule(lr){5-6} \cmidrule(lr){7-8}
        Scenario & Difference & Freq & \% & Freq & \% & Freq & \% \\
        \midrule
        \multirow{4}{*}{Scenario 1}
            & $-1$   &   29 &  0.94 &   90 &  2.92 &  499 & 16.20 \\
            & $0$    & 2960 & 96.10 & 2786 & 90.45 & 2569 & 83.41 \\
            & $1$    &   91 &  2.95 &  204 &  6.62 &   12 &  0.39 \\
            & Total  & 3080 & 100.00 & 3080 & 100.00 & 3080 & 100.00 \\
        \midrule
        \multirow{4}{*}{Scenario 2}
            & $-1$   &   17 &  0.62 &   32 &  1.17 &    134 &  4.91 \\
            & $0$    & 2675 & 98.06 & 2627 & 96.30 &    2592 &  95.2 \\
            & $1$    &   36 &  1.32 &   69 &  2.53 &      2 &  0.07 \\
            & Total  & 2728 & 100.00 & 2728 & 100.00 &    2728 & 100.00 \\
        \bottomrule
    \end{tabular}
\end{table}
	In both Scenarios 1 and 2, \texttt{BBMF} and  \texttt{Asso} produce substantially fewer errors than \texttt{GreConD+}. 
    
    \subsection{Clustering of the Matrices}
    \label{subsec:clust_sim}
\begin{figure}[htbp]
        \centering
        %--- Subfigure (a) ---
        \begin{subfigure}{\textwidth}
		      \centering
		      \includegraphics[width=\textwidth]{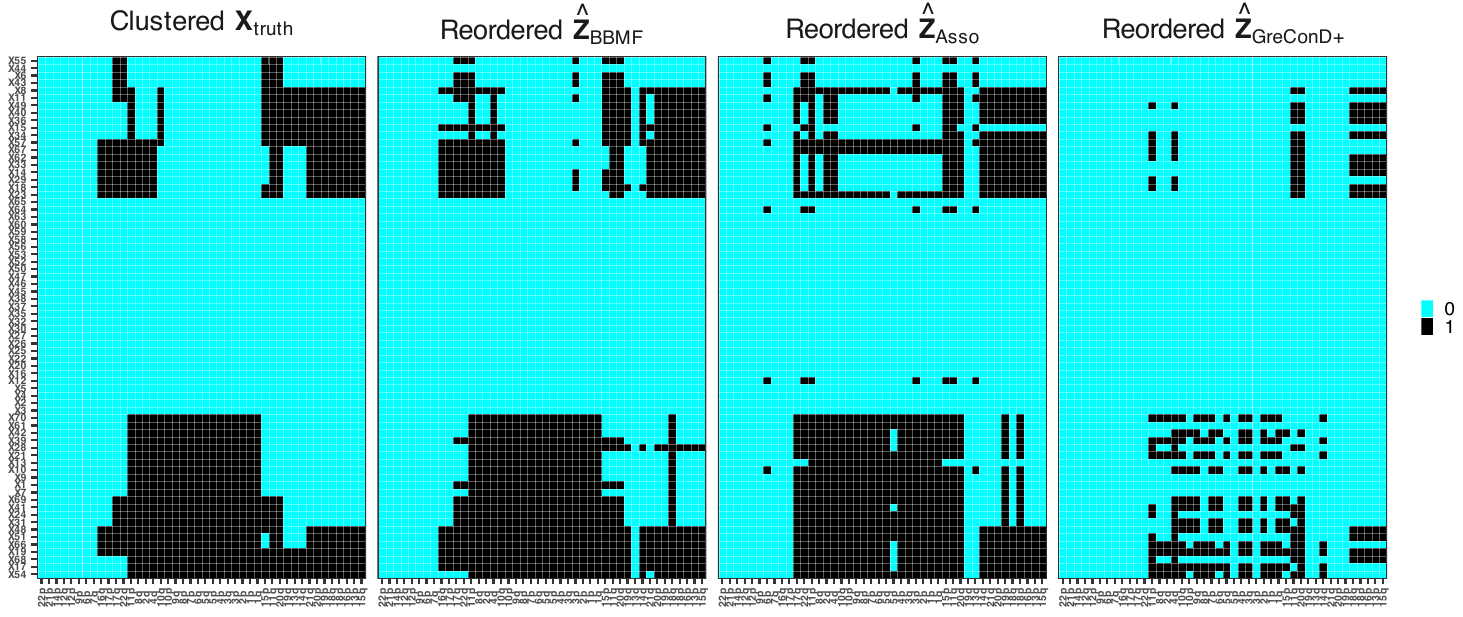} 
		      \caption{Scenario 1.}
		  \label{fig:clustered_Sim_Expt1}
    \end{subfigure}
    
	\vspace{0.5em}
    %--- Subfigure (b) ---
    \begin{subfigure}{\textwidth}
		\centering
		\includegraphics[width=\textwidth]{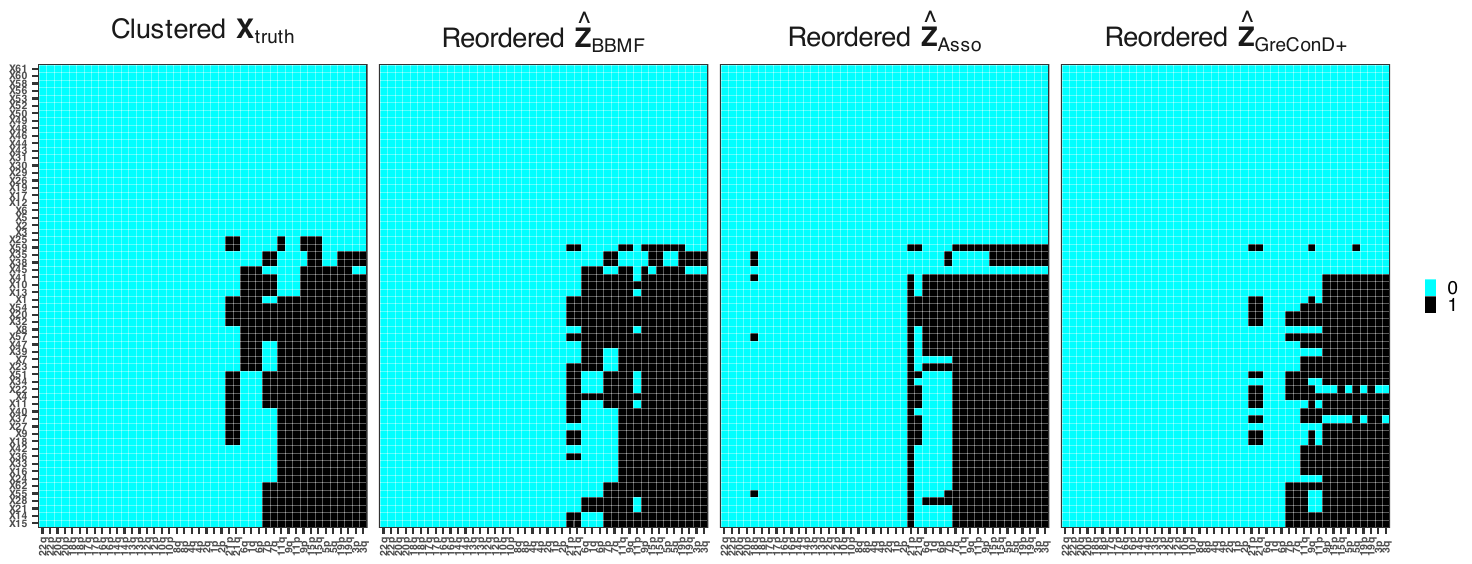}
		\caption{Scenario 2.}
		\label{fig:clustered_matrices}
         \end{subfigure}
         \caption{Clustered ground-truth matrix and reordered reconstructions obtained from \texttt{BBMF}, \texttt{Asso} and \texttt{GreConD+}.}
         \label{fig:clustered_matrices_all}
	\end{figure}

    To visualize reconstruction quality, we first applied hierarchical clustering to the rows and columns of the ground-truth matrix $\mathbf{X}_{\text{truth}}$. Distances between binary profiles were computed using cosine distance, $d_{\text{cos}} = 1 - \text{cosine similarity}$, and clusters were obtained by complete-linkage agglomerative clustering. The reconstructed matrices $\hat{\mathbf{Z}}_{\text{BBMF}}$, $\hat{\mathbf{Z}}_{\text{Asso}}$, and $\hat{\mathbf{Z}}_{\text{GreConD+}}$, as well as the difference heatmaps in Figure~\ref{fig:clustered_matrices_all}, were reordered using these same row and column orderings so that discrepancies are directly comparable and aligned with the latent block structure of $\mathbf{X}_{\text{truth}}$.
    
	In both scenarios, the reordered representation $\hat{\mathbf{Z}}_{\text{BBMF}}$ aligns closely with the clustered ground-truth matrix $\mathbf{X}_{\text{truth}}$: the main block of 1s appears in the correct location and contains very few spurious entries, indicating that the underlying block structure is faithfully reproduced. The reordered $\hat{\mathbf{Z}}_{\text{Asso}}$ exhibits more scattered ones outside the main block, as well as missing ones within it while $\hat{\mathbf{Z}}_{\text{GreConD+}}$ displays large contiguous regions of mismatch. The clustered ordering therefore reinforces the conclusion that \texttt{BBMF} recovers the true block structure more accurately than \texttt{Asso} and \texttt{GreConD+}.

	\subsection{BBMF Estimated Bicliques Decomposition for Scenario 2}
\begin{figure}[htbp]
		\centering
		\includegraphics[width=\textwidth]{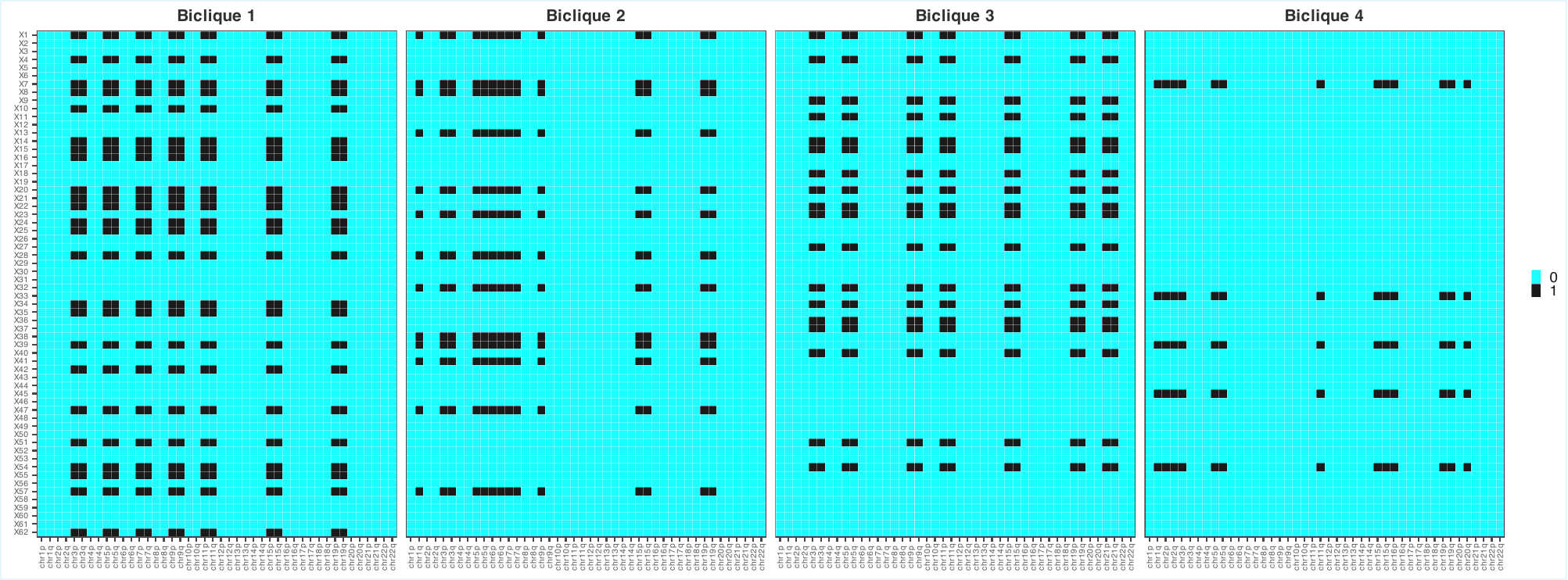}
		\caption{Visualization of the Bicliques resulting from the application of \texttt{BBMF} of the Synthetic Data.}
		\label{fig:synth_X1_Bicliques}
        %\textcolor{red}{TODO: panel labels in this figure likely use $\hat{\tilde{X}}_{\text{BBMF}}$ from the original notation; regenerate with $\hat{Z}_{\text{BBMF}}$ to match the manuscript.}
	\end{figure}

	Following Proposition~\ref{prop:bicliq_mat}, we construct the bicliques associated with the MAP factorization $(\hat{\mathbf{W}}_{\text{BBMF}}, \hat{\mathbf{H}}_{\text{BBMF}})$; combined via OR they reproduce $\hat{\mathbf{Z}}_{\text{BBMF}}$,exactly and through it approximate the ground-truth matrix $\mathbf{X}_{\text{truth}}$. The biclique heatmaps are presented in Figure~\ref{fig:synth_X1_Bicliques}.
	
	Bicliques 1 and 3 contain moderately dense patches of 1’s across several contiguous chromosomal arms, shared by a consistent group of patients. Biclique 2 shows a more localized pattern, with compact clusters of 1’s in fewer chromosomal regions but aligned across many patients. Biclique 4 is sparser, with isolated pockets of 1’s suggesting a weaker or more selective signal.

    By analogy to the real-data biology examined in Section~\ref{sec:applic_real_data}, these biclique patterns are reminiscent of distinct modes of genomic alteration: Bicliques~1 and~3, with broad, dense activations over contiguous chromosomal arms, resemble recurrent arm-level copy-number gains or losses shared by many tumours; Biclique~2, with compact clusters at fewer loci, resembles more focal events; and the sparser, fragmented pattern in Biclique~4 resembles rare or heterogeneous alterations in smaller patient subsets. As these are synthetic data, the resemblance reflects the patterns embedded in the simulation rather than independent biological discovery.
    
	\section{Applications to Real Data}
	\label{sec:applic_real_data}

     The \texttt{BBMF} model is applied to binary CNA data from multiple myeloma patients, originally collected and analyzed in \citet{samur2023high}, who studied the mutational effects of high-dose melphalan (HDM) on surviving myeloma cells using paired samples at diagnosis and relapse. The present analysis uses only the diagnostic samples. The data were discretised using the methodology of \citet{shen2016facets}: if the average copy number within an arm is above $2.2$ the arm is coded as ``gain'' ($1$); if below $1.8$, ``deletion'' ($-1$); the remainder, in the range $1.8$--$2.2$, ``normal'' ($0$). The resulting ternary matrix is then binarised by mapping ``gain'' to amplification ($1$) and ``normal''/``deletion'' to non-amplification ($0$); we revisit the implications of discarding the deletion information in Section~\ref{sec:summary_conclusions}. The data consist of $62$ patient-samples and $44$ chromosomal arms.
     
	\subsection{BBMF MAP Estimates for the CNA Data}

\begin{figure}[htbp]
		\centering
		\resizebox{\textwidth}{!}{%
			\begin{tabular}{cccccc}
				\adjustbox{valign=m}{\includegraphics[height=6cm,width=.95cm]{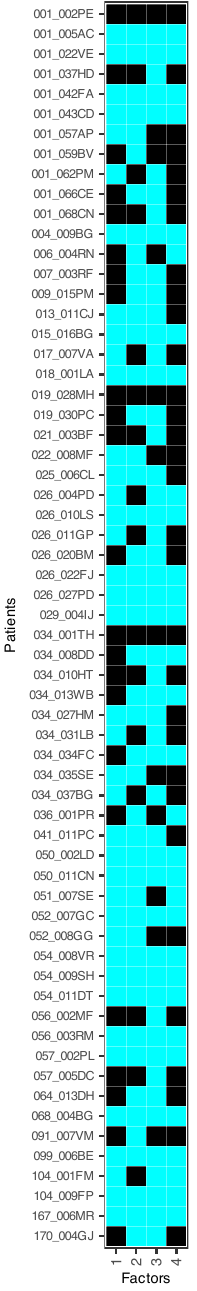}} &
				\adjustbox{valign=m}{$\circ$} &
				\adjustbox{valign=m}{\includegraphics[height=.85cm,width=6cm]{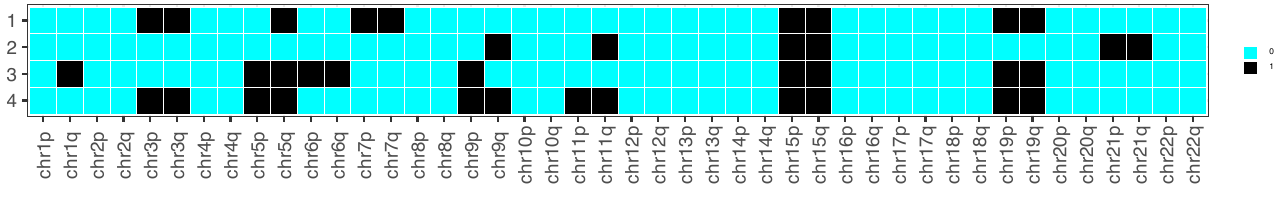}}&
				\adjustbox{valign=m}{$=$}&
				\adjustbox{valign=m}{\includegraphics[height=6cm,width=6cm]{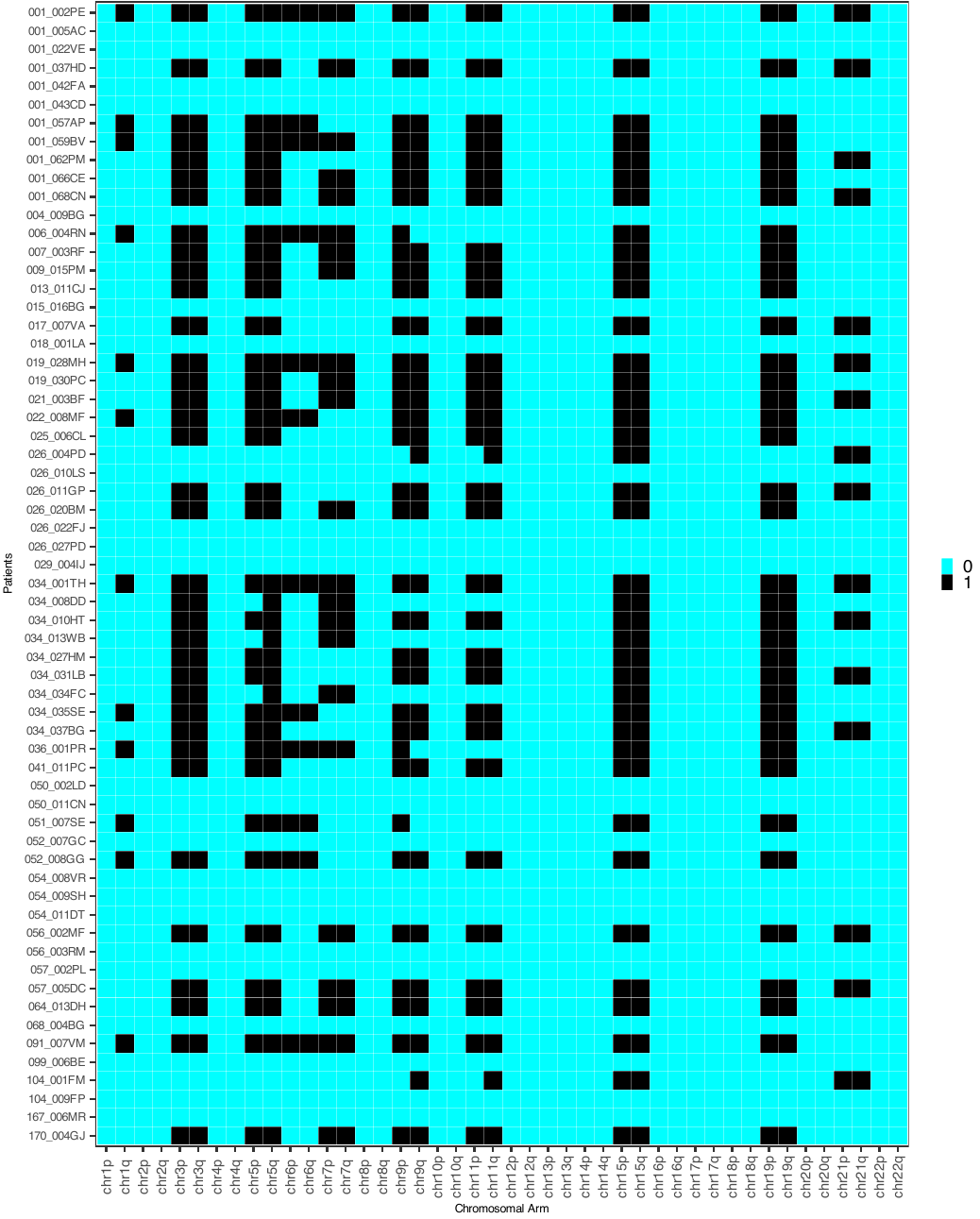}} \\
				$\hat{\mathbf{W}}$  &  &$\hat{\mathbf{H}}$&& $\hat{\mathbf{Z}}$
			\end{tabular}
		}
		\caption{MAP estimates of the factor matrices for the CNA data.}
		\label{fig:X1_Diag_BBMF}
	\end{figure}

	To infer the latent structure in the observed Boolean data, we applied the proposed \texttt{BBMF} model via the Gibbs sampler of Algorithm~\ref{alg:gibbs}, running four independent Markov chains with $R = 4$ latent factors for $100{,}000$ iterations each. We specified independent $\mathrm{Beta}(1,1)$ hyperpriors for all beta-distributed parameters, i.e., $(a_1,a_2) = (b_1,b_2) = (c_1,c_2) = (d_1,d_2) = (1,1)$, so that each corresponding parameter has a uniform prior on $[0,1]$. This choice is weakly informative and non-directional, representing lack of prior information about the parameter values and avoiding arbitrary asymmetries across components. For posterior inference, we discarded the first $20{,}000$ iterations of each Markov chain as burn-in. We monitored the log-likelihood, log-posterior, and log-prior traces across iterations and observed agreement across chains; the trace plots are reported in Figure~S1 of the Supplementary materials. The MAP estimates obtained per Section~\ref{subsec:inference} are presented in Figure~\ref{fig:X1_Diag_BBMF}. 
	%--------------------------------------------------------------------------------
	%--------------------------------------------------------------------------------
    In this factorization, each row of $\hat{\mathbf{W}}$ corresponds to a patient and each column to a latent factor, with a $1$ indicating that the corresponding factor pattern encoded in $\hat{\mathbf{H}}$ is active for that patient. The chromosomal amplification profile of a patient is thus the Boolean OR of all factor patterns active in their row of $\hat{\mathbf{W}}$, while factors with zero entries contribute nothing to the profile. These binary activation patterns may delineate clinically or biologically meaningful subgroups---such as hyperdiploid and non-hyperdiploid MM---or stratify patients according to high-risk genetic lesions, highlighting the heterogeneity of chromosomal signatures in MM.

    Each row of $\hat{\mathbf{H}}$ corresponds to a factor and each column to a chromosomal arm, where a $1$ indicates that amplification of that arm is a defining feature of the factor. The factor recovered by \texttt{BBMF} aligns with the classic hyperdiploid MM signature reported in the literature \citep{ankathil2021hyperdiploid, chretien2015understanding}, characterised by trisomies of chromosomes 3, 5, 7, 9, 11, 15, 19, and 21 (i.e.\ amplification of both arms: chr3p, chr3q, chr5p, chr5q, chr7p, chr7q, chr9p, chr9q, chr11p, chr11q, chr15p, chr15q, chr19p, chr19q, chr21p, chr21q). The recovered factors additionally identify the separate, secondary high-risk amplification of 1q (chr1q) \citep{binder2017prognostic} and the absence of 13q amplification (consistent with 13q being typically lost rather than gained in MM).

\begin{figure}[htbp]
		      \centering
		      \includegraphics[width=.9\textwidth]{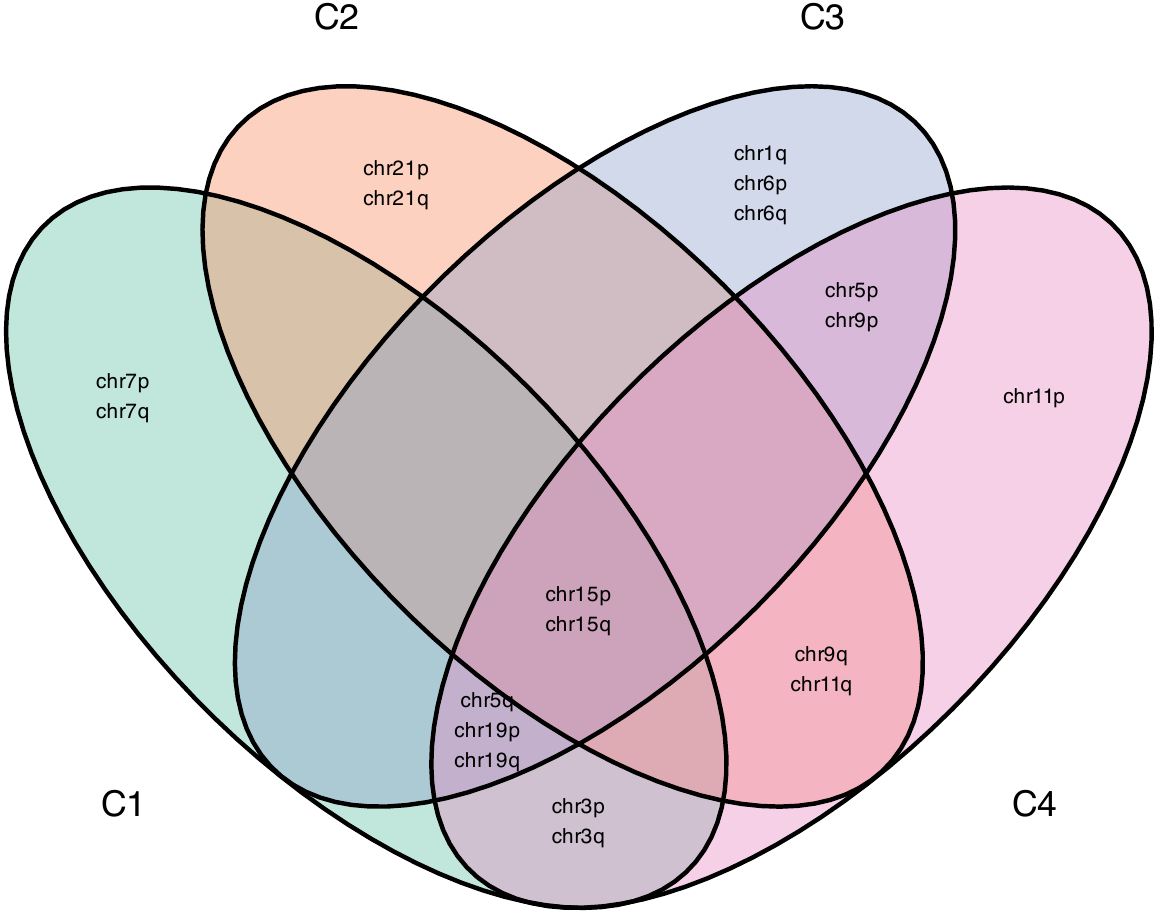}
		      \caption{Chromosomal arms associated with classical hyperdiploidy in multiple myeloma, as identified in the bicliques (C1--C4). Each biclique represents a co-amplified subset of chromosomal arms; the diagram shows the intersections and exclusive contributions of each biclique.}
		      \label{fig:VennBBMF}
	   \end{figure}

    A summary of the identified chromosomal arms is presented in the Venn diagram in Figure~\ref{fig:VennBBMF}.
	There is an overlap among bicliques revealing several chromosomal arms (e.g., chr3p, chr3q, chr19p, chr19q) that are shared by all four patterns. Notably, chr7p and chr7q are unique to C1, while chr5p, chr5q, chr9p, chr9q, and chr11p are exclusive to C4, indicating specific subgroups of co-amplifications.
	
	\subsection{Cluster-Informed Comparison of Original and BBMF-Reconstructed CNA Matrices}
\begin{figure}[htbp]
		\centering
		\includegraphics[width=\textwidth]{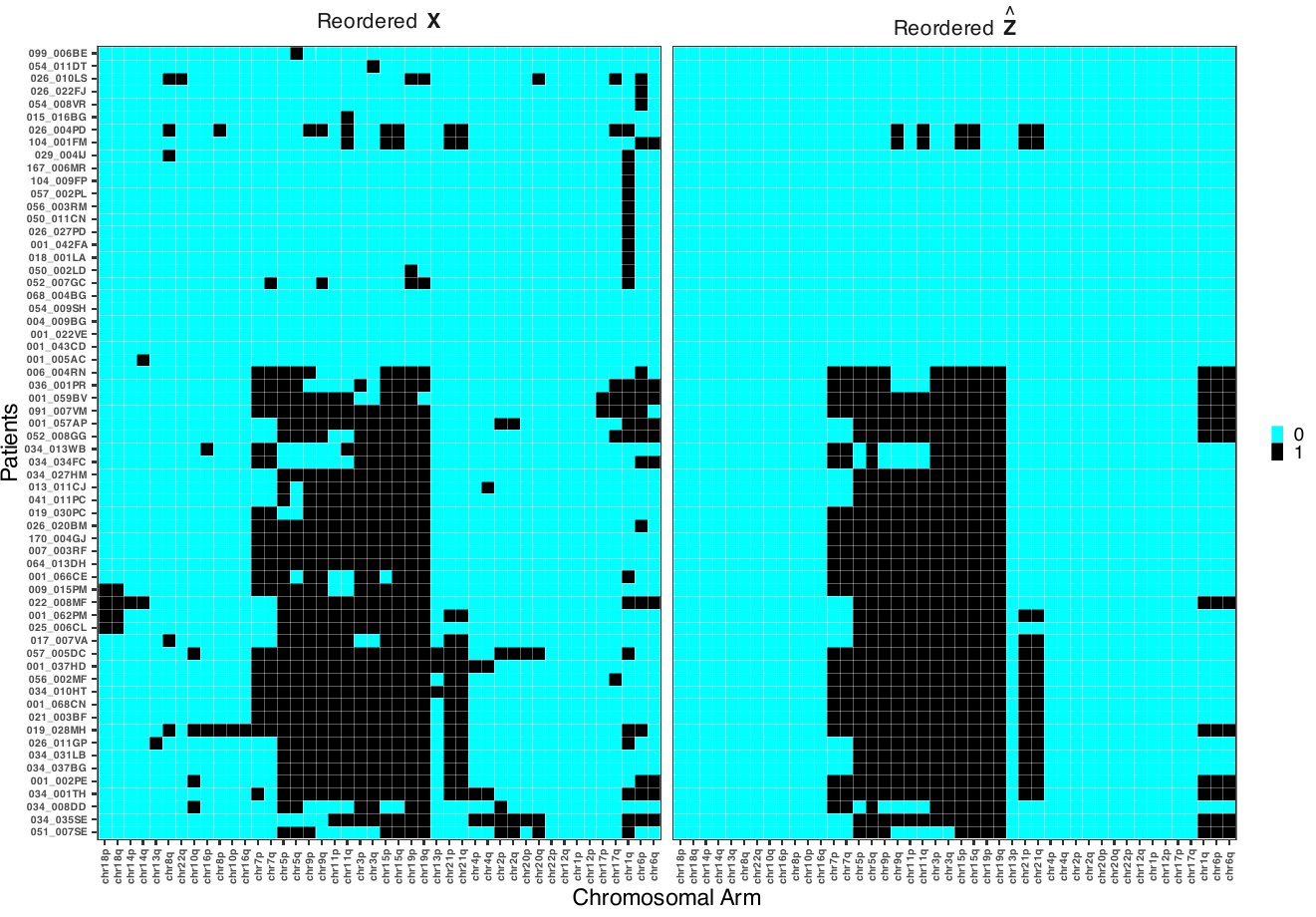}
		\caption{Comparison of the original and BBMF-reconstructed CNA matrices for the multiple myeloma data, visualised with cluster-informed ordering.}
		\label{fig:X1_BBMF_ordered}
	\end{figure}

	To compare the original and reconstructed chromosomal-arm alteration matrices, we first computed an ordering of patients (rows) and chromosomal arms (columns) from the original matrix using hierarchical clustering on a cosine distance (i.e., $1 - \cos\theta$ on the binarised vectors) as previously described in Section~\ref{subsec:clust_sim}. We then applied the resulting row and column permutations to both the original matrix and each reconstruction before visualization. This ensures that the same patients and chromosomal arms appear in the same positions across panels, so differences can be interpreted directly. The resulting heatmaps are shown in Figure~\ref{fig:X1_BBMF_ordered}.

    Figure~\ref{fig:X1_BBMF_ordered} shows that the \texttt{BBMF} reconstruction appears to capture the main structure of the original data: distinct blocks of binary amplification, representing major patient clusters and co-amplified chromosomal regions, are visible in both matrices, suggesting that the factorization recovers the dominant patterns of variation and co-occurrence among amplifications. Unlike the simulation experiments of Section~\ref{sec:sim_Expts}, no gold-standard $\mathbf{X}_{\text{truth}}$ is available for the real CNA data, so this assessment is necessarily qualitative; inspecting the reconstruction visually remains informative for judging whether the recovered structure is plausible in a real problem.

    Some differences are also apparent. In regions where the original data display more heterogeneous or sparse patterns, the reconstruction either smooths cluster boundaries or fails to reproduce isolated alteration events, reflecting the inherent trade-off between denoising and fidelity, limitations in resolving fine-scale structure, or genuine shifts in chromosomal alteration patterns due to disease evolution.

	\subsection{BBMF Estimated Bicliques Decomposition for the CNA Data}
\begin{figure}[tbp]
		\includegraphics[width=\textwidth]{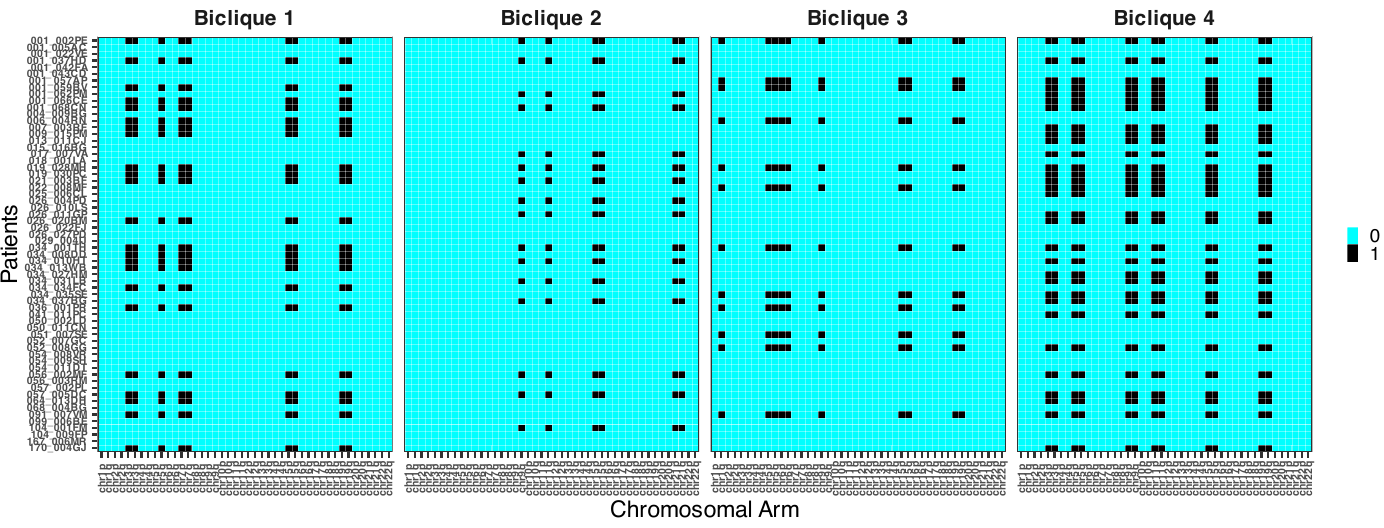}
		\caption{Decomposition of the CNA data into bicliques using the BBMF model, revealing latent factors in chromosomal alteration in multiple myeloma.}
		\label{fig:diagnostic_bicliques}
	\end{figure}

    Based on previously estimated factor matrices $\mathbf{W}$ and $\mathbf{H}$, we constructed bicliques using Proposition~\ref{prop:bicliq_mat}. Each resulting biclique captures a group of patients that exhibit a coherent alteration profile on a specific set of chromosomal arms. The results are presented in Figure~\ref{fig:diagnostic_bicliques}. 
	Within each biclique, a coherent block of 1s links a subset of patients to a characteristic set of chromosomal arms, indicating groups of loci that tend to be altered together. Different subsets of patients are associated with distinct sets of arms, suggesting molecular subgroups defined by their CNA signatures.Several chromosomal arms participate in more than one biclique, pointing to shared ``hotspot'' regions that co-occur with multiple CNA modules.
	%--------------------------------------------------------------------------------------------
	\section{Summary and Conclusions}
	\label{sec:summary_conclusions}
	We have introduced a Bayesian Boolean Matrix Factorization (\texttt{BBMF}) model with three contributions relative to prior probabilistic BooMF: an asymmetric two-parameter noise model that decouples sensitivity from the false-positive rate; a hierarchical Beta with continuous spike-and-slab prior on factor-activation probabilities, which lets the data prune the effective factorization rank below the user-supplied upper bound $R$; and an aggregate factor-alignment similarity (AFAS) diagnostic for monitoring chain-level factor-recovery stability. We benchmarked \texttt{BBMF} against the established \texttt{Asso} and \texttt{GreConD+} algorithms on two simulation scenarios, then applied it to arm-level copy-number alteration data from multiple myeloma.
	
	For factor recovery, \texttt{BBMF}  produced similarity matrices for ground truth $\mathbf{H}_{\text{truth}}$ versus $\hat{\mathbf{H}}_{\text{BBMF}}$, which were strongly diagonally dominant for most true factors, with low off–diagonal similarity, indicating well-separated latent components. By contrast, \texttt{Asso} tended to recover only one factor almost perfectly and showed substantial mixing among the remaining factors, with several true components sharing comparable similarity to multiple estimated factors. Similarly, the \texttt{GreConD+} algorithm generally shows poorer latent factor recovery as well as higher reconstruction error. Visual inspection of the factor matrices confirmed this pattern: \texttt{Asso} captured some broad activation structure but introduced noticeable spurious activations and omissions, whereas  \texttt{BBMF} yielded factor matrices that, once reordered, closely mirrored the true binary patterns with relatively few spurious entries.
	
	Reconstruction accuracy yields the same ranking. In both Scenario~1 and Scenario~2, \texttt{BBMF} attains the highest F1 and MCC and the lowest reconstruction error rate (Table~\ref{tab:combined_metrics}); \texttt{GreConD+}'s highest specificity comes at the cost of substantially lower F1/MCC, reflecting an overly conservative bias toward predicting zeros. The gap between \texttt{BBMF} and the heuristic baselines is most pronounced in Scenario~1.
	
	Finally, we applied \texttt{BBMF} to copy-number alteration data from multiple myeloma patients. The model identified a few interpretable latent bicliques linking subsets of patients to chromosomal arms with recurrent co-alteration. The reconstructed clustered matrices preserved the main block structures of the original data, suggesting that the factorised representations capture key biological axes of variation. A limitation of the present analysis is that the binarisation in Section~\ref{sec:applic_real_data} retains only amplifications: the original ternary CNA data also contain deletion events (e.g.\ 13q loss, 17p loss, 1p loss), which are biologically informative in MM but are pooled with ``normal'' in the binary encoding. Extending \texttt{BBMF} to handle ternary inputs --- either by jointly modelling separate amplification and deletion factor matrices, or by adopting a multi-class Boolean generalisation --- is a natural direction for future work. Even with this restriction, \texttt{BBMF} appears to be a practical, interpretable method for uncovering discrete latent structure in complex biomedical data.
	
	%%%%%%%%%%%%%%%%%%%%%%%%%%%%%%%%%%%%%%%%%%%%%%
	%% Acknowledgements                         %%
	%% should be provided in the                %%
	%% Acknowledgements section.                %%
	%%%%%%%%%%%%%%%%%%%%%%%%%%%%%%%%%%%%%%%%%%%%%%
	%\begin{acks}[Acknowledgments]
	%	The authors would like to thank the anonymous referees, an Associate
	%	Editor and the Editor for their constructive comments that improved the
	%	quality of this paper.
	%\end{acks}
	
	%%%%%%%%%%%%%%%%%%%%%%%%%%%%%%%%%%%%%%%%%%%%%%
	%% Funding information, if any,             %%
	%% should be provided in the                %%
	%% funding section.                         %%
	%%%%%%%%%%%%%%%%%%%%%%%%%%%%%%%%%%%%%%%%%%%%%%
	\begin{funding}
		This research was supported by a gift from Anand Krishnamurthy \& Ruth Lievano. 
	\end{funding}

	\bibliographystyle{ba}
	\let\enquote\undefined  % avoid clash with csquotes; ba.bst redefines \enquote in main.bbl
	\bibliography{Refs_BBMF,Refs_BBMF_added}
    \newpage
    
	%%%%%%%%%%%%%%%%%%%%%%%%%%%%%%%%%%%%%%%%%%%%%%
	%% Supplementary Material, including data   %%
	%% sets and code, should be provided in     %%
	%% {supplement} environment with title      %%
	%% and short description. It cannot be      %%
	%% available exclusively as external link.  %%
	%% All Supplementary Material must be       %%
	%% available to the reader on Project       %%
	%% Euclid with the published article.       %%
	%%%%%%%%%%%%%%%%%%%%%%%%%%%%%%%%%%%%%%%%%%%%%%
    \begin{supplement}
    \stitle{Appendixes for A Bayesian Boolean Matrix Factorization with Application to Copy Number Analysis in Cancer}
    \sdescription{The supplementary material provides additional  visualization that support the main manuscript.}
    \end{supplement}
    \end{document}

\begin{frontmatter}
		\title{Appendixes for A Bayesian Boolean Matrix Factorization with Application to Copy Number Analysis in Cancer}
		%\title{A sample article title with some additional note\thanksref{t1}}
		%\runtitle{A sample running head title}
		%\thankstext{T1}{A sample additional note to the title.}
		
		\begin{aug}
			%%%%%%%%%%%%%%%%%%%%%%%%%%%%%%%%%%%%%%%%%%%%%%%
			%% Authors                                  %%
			%%%%%%%%%%%%%%%%%%%%%%%%%%%%%%%%%%%%%%%%%%%%%%%
			\author[A,B]{\fnms{Adolphus }~\snm{Wagala}}
			\author[A,B]{\fnms{Mehmet}~\snm{Samur}}
			\and
			\author[A,B]{\fnms{Giovanni}~\snm{Parmigiani}}
			
			%%%%%%%%%%%%%%%%%%%%%%%%%%%%%%%%%%%%%%%%%%%%%%%
			%% Addresses                               %%
			%%%%%%%%%%%%%%%%%%%%%%%%%%%%%%%%%%%%%%%%%%%%%%%
			\address[A]{Department of Data Science, Dana-Farber Cancer Institute}
			
			\address[B]{Department of Biostatistics, Harvard T.H. Chan School of Public Health}
			
			\runauthor{Wagala, Samur and Parmigiani}
		\end{aug}
			\appendix
		\begin{abstract}
			This document contains the appendices for ``A Bayesian Boolean Matrix Factorization with Application to Copy Number Analysis in Cancer.'' It provides additional  visualization that support the main manuscript.
		\end{abstract}
		
	\end{frontmatter}
		\section{Trace Plots when BBMF is Applied to Simulated and Real Data}
		In our matrix factorization model $\mathbf{X} \approx \mathbf{W}\circ \mathbf{H}$, trace plots of the log-likelihood and log-prior provide only a necessary, but not sufficient, indication of convergence, since many distinct factorizations $(\mathbf{W,H})$ can yield similar log-density values. 
		\begin{figure}[H]
			\centering
			\includegraphics[width=\textwidth, height=0.4\textheight]{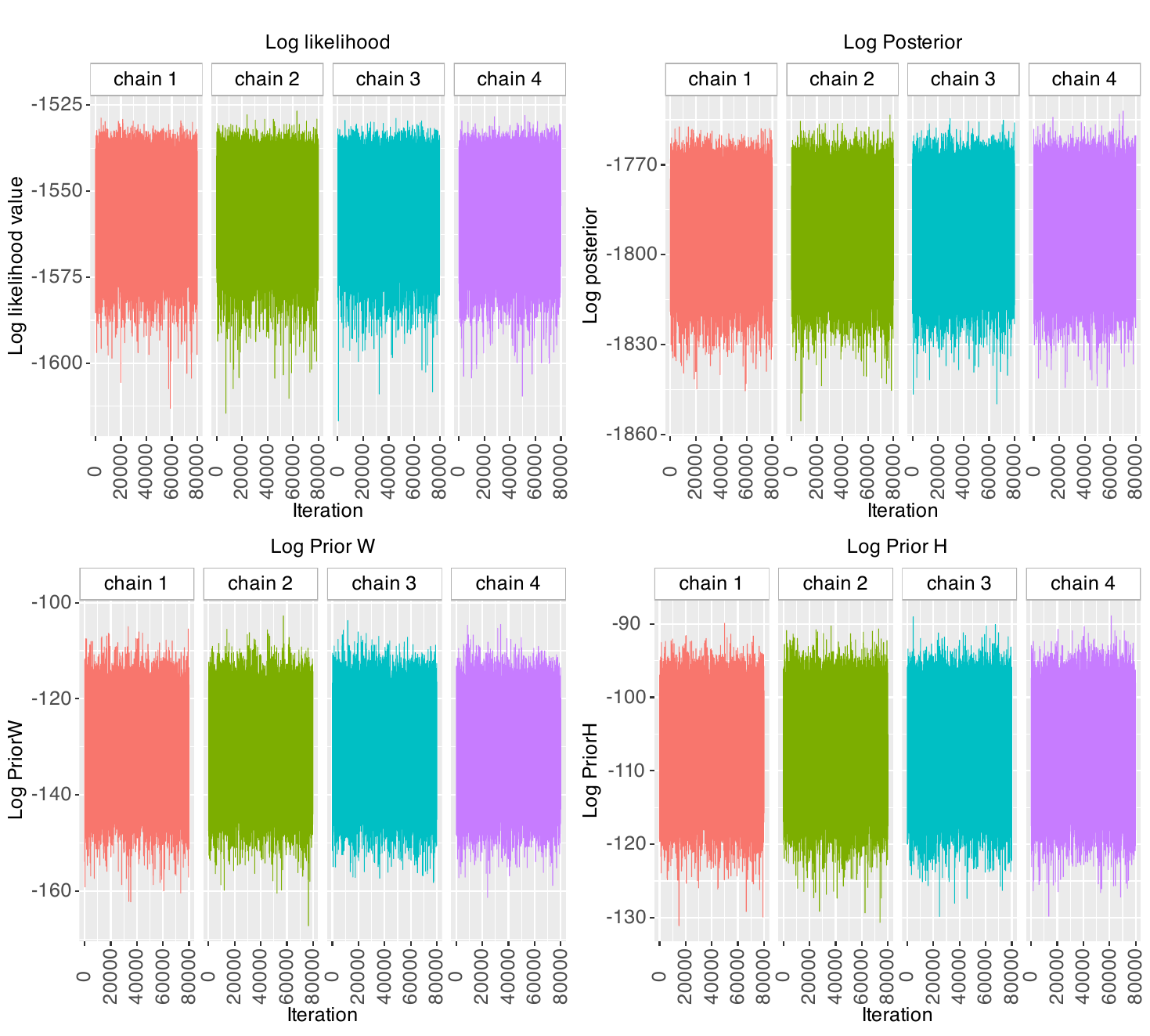}
			\caption{Trace plots of log-likelihood, log-posterior, and log-prior terms ($\mathbf{W}$ and $\mathbf{H}$) for four MCMC chains for the Scenerio 2 data after burn-in removal.}
			\label{fig:Trace_Plots_Sim_Expt1}
		\end{figure}
		%--------------------------------------------------------------------------------------
		\begin{figure}[H]
			\centering
			\includegraphics[width=\textwidth, height=0.4\textheight]{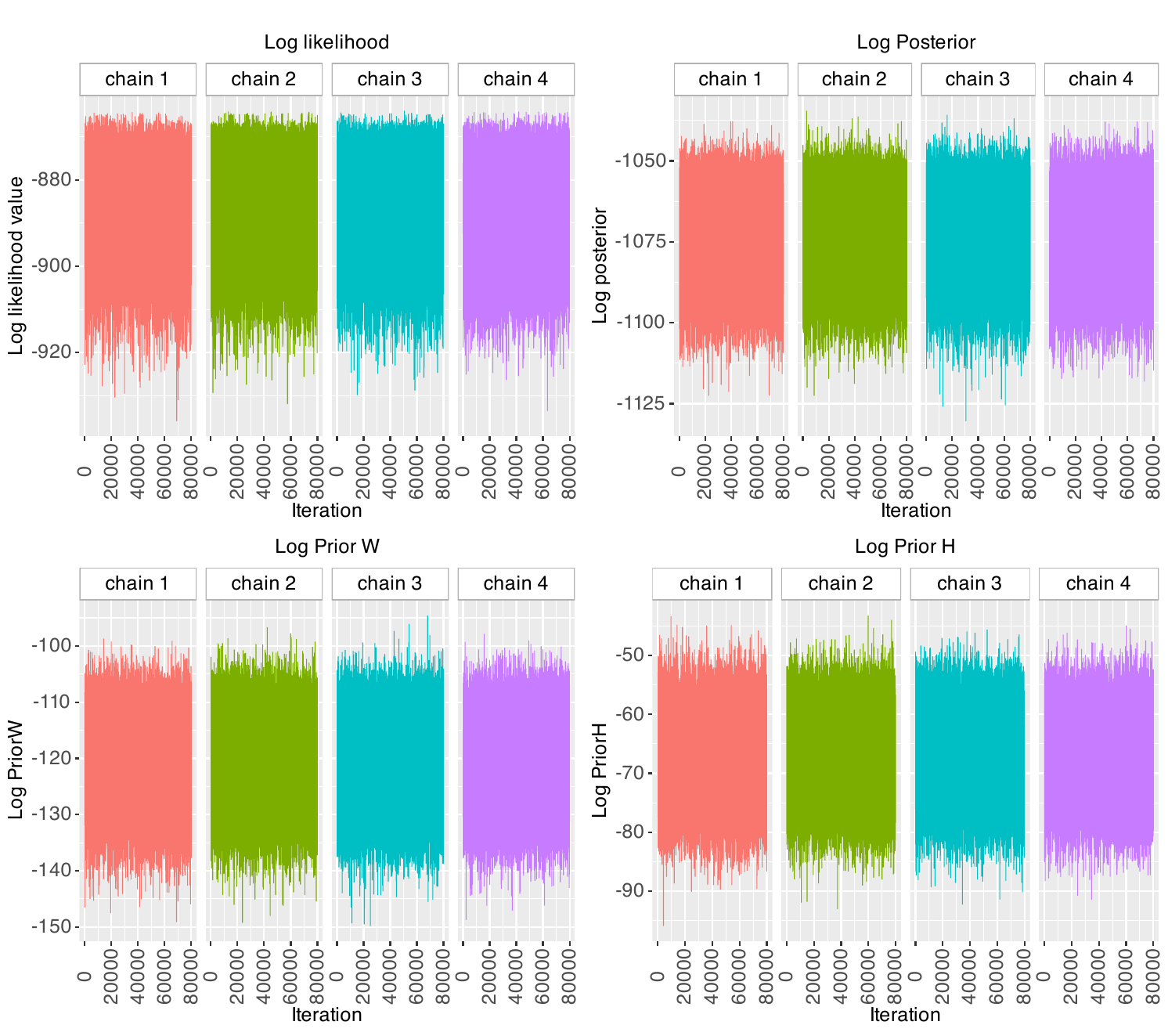}
			\caption{Trace plots of log-likelihood, log-posterior, and log-prior terms ($\mathbf{W}$ and $\mathbf{H}$) for four MCMC chains for the Scenerio 2 data after burn-in removal.}
			\label{fig:Trace_Plots_Sim_Expt2}
		\end{figure}
		%--------------------------------------------------------------------------------------
		\begin{figure}[H]
			\centering
			\includegraphics[width=\textwidth,height=0.4\textheight]{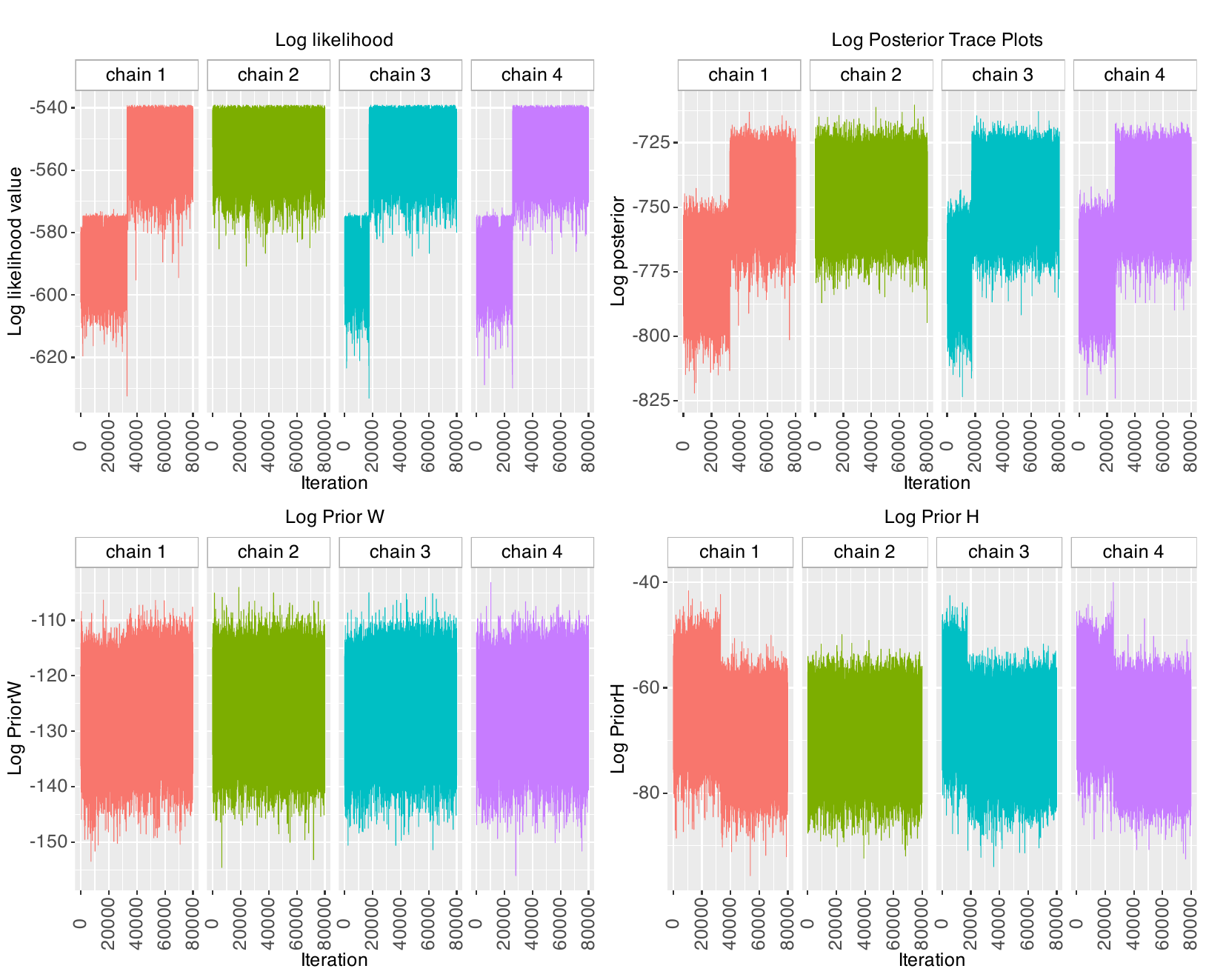}
			\caption{Trace plots of log-likelihood, log-posterior, and log-prior terms ($\mathbf{W}$ and $\mathbf{H}$) for four MCMC chains for the CNA data after burn-in removal.}
			\label{fig:Trace_Plots_CNA}
		\end{figure}